\pdfoutput=1

\documentclass[11pt]{article}

\usepackage[preprint]{acl}

\usepackage{times}
\usepackage{latexsym}

\usepackage[T1]{fontenc}

\usepackage[utf8]{inputenc}

\usepackage{microtype}

\usepackage{inconsolata}

\usepackage{graphicx}

\usepackage{amsmath}
\usepackage{enumitem}
\usepackage{booktabs}
\usepackage{multirow}
\usepackage{float}
\usepackage{subcaption}

\usepackage{colortbl}
\definecolor{green3}{HTML}{BFD8B6}
\definecolor{green2}{HTML}{E7F0E5}

\usepackage[disable]{todonotes}
\newcommand{\mydonetodo}[1]{}
\newcommand{\doneinlinetodo}[1]{}
\newcommand{\inlinetodo}[1]{}

\usepackage{xspace}
\newcommand{\llmagg}{\textsc{LLM-AggreFact}\xspace}
\newcommand{\bespoke}{Bespoke-7B\xspace}
\newcommand{\gptf}{gpt-4-turbo\xspace}
\newcommand{\gptt}{gpt-3.5-turbo\xspace}
\newcommand{\mft}{MiniCheck-FT5\xspace}
\newcommand{\mroberta}{MiniCheck-Rbta\xspace}
\newcommand{\ais}{AutoAIS\xspace}

\title{Verify with Caution:\\The Pitfalls of Relying on Imperfect Factuality Metrics}

\author{Ameya Godbole \and Robin Jia \\
  University of Southern California \\
  \texttt{\{ameyagod,robinjia\}@usc.edu} \\}

\begin{document}
\maketitle
\begin{abstract}
Improvements in large language models have led to increasing optimism that they can serve as reliable evaluators of natural language generation outputs. 
In this paper, we challenge this optimism by thoroughly re-evaluating five state-of-the-art factuality metrics on a collection of 11 datasets for summarization, retrieval-augmented generation, and question answering.
We find that these evaluators are inconsistent with each other and often misestimate system-level performance, both of which can lead to a variety of pitfalls.
We further show that these metrics exhibit biases against highly paraphrased outputs and outputs that draw upon faraway parts of the source documents.
We urge users of these factuality metrics to proceed with caution and manually validate the reliability of these metrics in their domain of interest before proceeding.
\end{abstract}

\section{Introduction}
\label{sec:intro}



  





\begin{figure}[t!]
    \centering
    \includegraphics[width=0.48\textwidth]{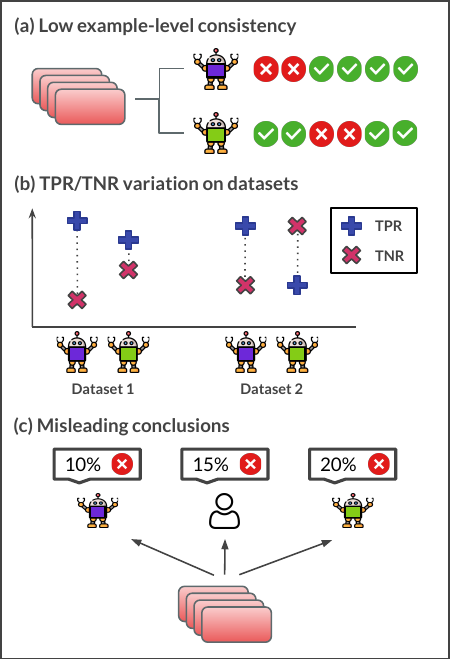}
    \caption{Selecting an AutoAIS evaluator based solely on balanced accuracy (BAcc) hides several underlying inconsistencies. Consider \gptf and \bespoke with comparable BAcc on \llmagg. The two evaluators have (a) low instance-level labeling consistency and (b) different true positive and true negative error rate trade-offs. (c) This results in different system-level evaluations when the evaluators are used downstream to evaluate the factuality of NLG systems. In several cases, one evaluator underestimates the human-labeled error rate while the other overestimates it.}
    \label{fig:eye-candy}
    \vspace{-3ex}
\end{figure}


Building automated evaluation metrics that match human judgment is difficult ongoing research~\citep{lambert2024rewardbenchevaluatingrewardmodels}. Past work has highlighted the flaws of automated evaluators in several NLP research domains, particularly machine translation~\citep[inter alia]{mathur-etal-2020-tangled,kocmi-etal-2021-ship}. Nonetheless, automated evaluation metrics are perennially appealing because they allow NLG system designers to bypass slower and costlier human evaluation.
Most recently, LLM-based metrics have led to optimism that NLG evaluation can be automated~\citep[inter alia]{kim-etal-2024-prometheus,vu-etal-2024-foundational}.
In particular, there is a growing demand for automated attribution evaluators, 
as LLMs are increasingly used for tasks in which factual reliability is crucial, such as summarization, retrieval-augmented generation, and open-ended chat \citep{gao-etal-2023-rarr,chen2023purrefficientlyeditinglanguage}.


In this work, we investigate automated metrics for the evaluation of ``Attribution to Identified Sources'' (AutoAIS; \citealp{rashkin-etal-2023-measuring}), i.e., judging whether a source document fully supports a claim.
We perform a comprehensive re-evaluation of 5 state-of-the-art \ais evaluators (2 proprietary and 3 open-source) on the \textbf{\llmagg} benchmark~\citep{tang2024minicheckefficientfactcheckingllms}, a collection of 11 datasets of claim-document pairs that are annotated for attributability.

We find several reasons to be cautious when using \ais evaluators.
First, state-of-the-art \ais evaluators with comparable leaderboard scores have large differences in predictions.
SotA evaluators have low agreement on an instance level (\S\ref{sec:f1-inconsistent}); error analysis based on different evaluators may yield different conclusions.
Evaluators can achieve comparable balanced accuracy by trading off true positive and true negative rates in different ways on different datasets (\S\ref{sec:f2-tpr-tnr-tradeoff}); evaluators cannot be relied on without verification on new datasets.
Second, evaluators also often give poor estimates of system-level error rate: \ais metrics on some datasets overestimate and on others underestimate how frequently unattributable claims are generated by a system (\S\ref{sec:f3-poor-system-error-est}).
This can lead to misestimation of the headroom for improvement on generation tasks (\S\ref{sec:f4-misleading-headroom}) and a poor ranking of systems (\S\ref{sec:f5-poor-system-ranking}); new system design ideas (such as new LMs, new decoding algorithms, etc) may be incorrectly cast aside based on imperfect automated metrics.


We identify 2 biases in the current SotA \ais metrics. In many domains, AutoAIS metrics struggle to detect unattributable claims with a high surface-level similarity with the document (\S\ref{sec:b1-rouge-bias}).
We also show that the performance of evaluators that chunk long reference documents is inherently limited because certain claims become unverifiable (\S\ref{sec:b2-chunking-bias}).
Both these properties---paraphrasing without directly copying
and synthesizing information from different parts of a long input document---are desirable in an NLG system and may be penalized if not appropriately addressed by evaluators.

In \S~\ref{sec:adjustment}, we attempt to reduce the bias/discrepancy between the labeled and predicted (estimated) system error rates. Threshold tuning to minimize the absolute bias on a calibration set is a consistent method for achieving low absolute bias.
For \ais evaluators that do not have a tunable threshold, posthoc adjustment of the estimated error rate \citep{10.1145/3117807-review-on-quantification} can reduce the absolute estimation bias (with certain caveats).


Finally, in \S\ref{sec:discussion}, we discuss the impact of these findings on downstream users of the \ais metrics, such as dataset developers and researchers studying how to improve the factuality of NLG systems. 
Since metrics do not yet transfer consistently to new datasets, we urge users of these metrics to first perform human validation of metric predictions in new data domains and on new systems. Finally, we urge developers of new \ais metrics to report a breakdown of metric behavior on the different error types across different bias axes of the evaluation data and with an evaluation of system-level error quantification.
\section{Problem Setup}

\subsection{Notation}
\label{sec:notation}

Given a claim $c$ and a document $d$\footnote{The document may be a composite of multiple evidence passages e.g. LFQA~\citep{chen2023understandingretrievalaugmentationlongform}.}, the role of the AutoAIS evaluator $\mathcal{A}$ is to judge whether all the information in $c$ is fully supported by the document $d$.\footnote{This is part of the definition of AIS given by \citet{rashkin-etal-2023-measuring}. Most AutoAIS systems assume decontextualization as a separate preprocessing step.} Following \citet{tang2024minicheckefficientfactcheckingllms}, we threshold the output of the evaluator at 0.5 and predict a label 0 (unattributable) or 1 (attributable). We will discuss the impact of tuning the threshold for downstream applications in \S~\ref{sec:adjustment}.
\begin{equation*}
    \setlength\abovedisplayskip{5pt}
    \setlength\belowdisplayskip{5pt}
    \mathcal{A}(d, c) \rightarrow \{0, 1\}
\end{equation*}
Certain \ais evaluators may have input length limits, in which case the document $d$ is segmented into chunks (of complete sentences) of a certain length $\{d^{(1)}, d^{(2)}, .., d^{(K)}\}$. Then the prediction:
\begin{equation*}
    \setlength\abovedisplayskip{5pt}
    \setlength\belowdisplayskip{6pt}
    \mathcal{A}(d, c) = \max_{k \in [1,K]} \mathcal{A}(d^{(k)}, c) \rightarrow \{0, 1\}
\end{equation*}



Our analysis will focus on the validation set of the \llmagg benchmark \citep{tang-etal-2024-minicheck}; a collection of 11 datasets with human-annotated attributability annotations.
We further split the examples from the RAGTruth dataset \citep{niu-etal-2024-ragtruth} in the benchmark into the 4 original subsets since they have qualitatively different inputs and task types. This results in a benchmark with 14 datasets.

Except for Wice and FactCheck-GPT, 12 of the 14 datasets contain generations from multiple systems. We use this to analyze the system-level error estimation and ranking of the different \ais evaluators. Appendix~\ref{app:dataset-details} provides a detailed breakdown of the datasets in the benchmark.

The benchmark assumes that each sentence is a standalone claim\footnote{\citet{tang-etal-2024-minicheck} showed that decontextualization and decomposition showed little improvement in the performance of the \ais evaluators.}\footnote{AggreFact-CNN treats the entire summary (avg of 3.2 sentences) as the claim because the dataset lacks sentence-level annotation.}. Except for AggreFact-CNN and AggreFact-XSum, 10 of the 12 datasets (with generations from multiple systems) originally contained multi-sentence responses that have been broken down into sentence-level examples in the benchmark. We evaluate the response-level performance of the \ais evaluators by mapping individual claims back to the original complete response.
We obtain a response-level factuality label by aggregating the claim-level labels. We adopt the strict definition of an attributable response~\citep{tang-etal-2024-tofueval}: a response is attributable if ALL claims in the response are attributable.

\subsection{\ais Evaluators}
\label{sec:metric-details}

The \llmagg benchmark ranks metrics based on average balanced accuracy (BAcc) across all data sets. BAcc of an evaluator is defined as the average of its True Positive Rate (TPR) and True Negative Rate (TNR) on a dataset, i.e. it measures the average performance of detecting the attributable and unattributable examples.

In this work, we study five evaluators from the \llmagg leaderboard. We choose 2 closed, API-based evaluators: \gptf \citep{openai2024gpt4technicalreport}(in particular, \texttt{gpt-4-0125-preview}) and \gptt (in particular, \texttt{gpt-3.5-turbo-0125}), and 3 open-weight models from the MiniCheck series~\citep{tang-etal-2024-minicheck}: Bespoke-Minicheck-7B (\bespoke), MiniCheck-FlanT5-Large (\mft) and MiniCheck-RoBERTa-Large (\mroberta). \bespoke and \gptf were the top evaluators on the leaderboard at the time of release. Similarly, \mft, \mroberta, and \gptt have very similar performances regarding average balanced accuracy across the datasets.

Evaluators with input length constraints (e.g. \mft, \mroberta, TRUE~\citep{honovich-etal-2022-true}, inter alia.) need to chunk the input documents to fit their max context window. To isolate the effect of chunking, we evaluate the \bespoke metric with chunked documents and compare the predictions against the original predictions without document chunking. In particular, we run the \bespoke metric as if it had a context window of 500 document tokens (same as \mft). We will refer to this setting as '\bespoke (cs=500)'.

\section{Re-Evaluating Factuality Metrics}
\label{sec:findings}

\subsection{Metrics have low consistency}
\label{sec:f1-inconsistent}

To study consistency between evaluation metrics, we measure the intersection-over-union (IoU) of the set of examples predicted as "unattributable" by the evaluators.
We find that for the two top-performing evaluators with similar balanced accuracy, \bespoke (Avg BAcc=77.4\%) and \gptf (Avg BAcc=76.2\%), the IoU is less than 50\% on 5 of the 14 datasets and less than 65\% on 9 of 14 datasets. The consistency is worse on the nine datasets where "unattributable" is the minority class (less than 25\% of the dataset). Refer to Appendix~\ref{app:metric-consistency} for the pairwise inconsistency of the 5 evaluation metrics studied.

This inconsistency has several implications. When scoring NLG systems, different evaluators may rank NLG systems differently and for different subsets of system predictions. We discuss this further in the next few sections. When conducting error analysis for NLG system development, different evaluators will highlight different "erroneous" unattributable examples. Using a single evaluator may highlight a biased subset of errors. We discuss this further in \S~\ref{sec:d2-benchmark-dev}.


\begin{figure*}[t]
    \centering
    \includegraphics[width=0.98\textwidth]{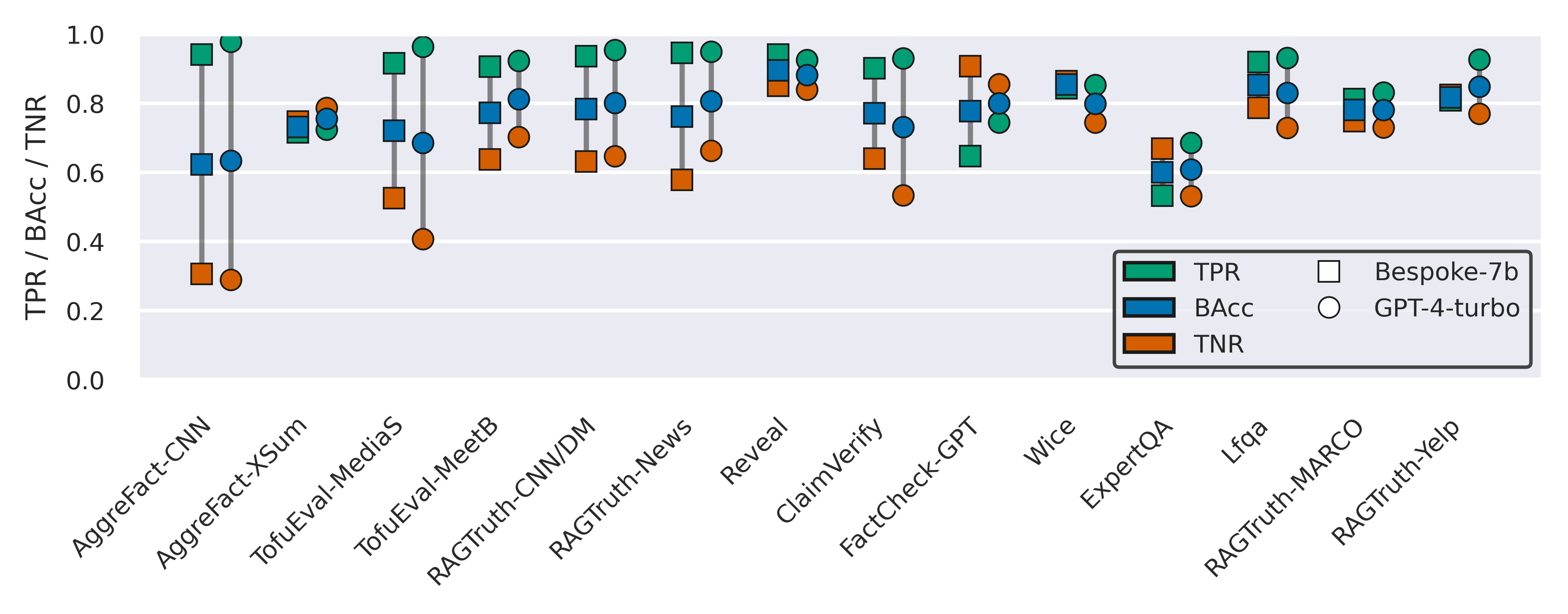}
    \vspace{-1em}
    \caption{\textbf{TPR/TNR/BAcc of evaluators across datasets.} Visualizing the breakdown of BAcc shows that \ais evaluators can have a large gap between TPR and TNR. Moreover, evaluators with the same BAcc can have different TPR and TNR trade-offs. In the extreme case of ExpertQA, GPT-4-turbo has a TPR of 68\% and TNR of 53\%, while Bespoke-7B has nearly the opposite performance.
    }
    \label{fig:tpr-tnr-trend}
    \vspace{-2ex}
\end{figure*}

\subsection{BAcc hides TPR/TNR trade-off}
\label{sec:f2-tpr-tnr-tradeoff}

Using balanced accuracy to evaluate \ais metrics hides the underlying trade-off between true-positive and true-negative rates. From Figure~\ref{fig:tpr-tnr-trend}, we see that the true positive and true negative rates for each evaluator vary widely across the datasets.\footnote{We report the false positive and false negative rates of the larger set of evaluation metrics in Tables~\ref{tab:full-fpr} and~\ref{tab:full-fnr}.} 
The gap between TPR and TNR is greater than 20\% on 7 of 14 datasets for \bespoke and \gptf.
By trading off TPR for TNR differently, different evaluators can achieve the same balanced accuracy. For example, on the FactCheck-GPT dataset, \bespoke achieves a BAcc of 77.7\% with a difference between TNR and TPR of 26\%. \gptf achieves a comparable BAcc of 80\% but with only an 11\% gap between TNR and TPR. Similarly and more surprisingly, \bespoke and \gptf achieve the same BAcc on the ExpertQA dataset but with inversed values of TPR and TNR.  

The trade-off between TPR and TNR has different implications for downstream users of the metric where the cost of type I and type II errors differs. We recommend metric designers report a breakdown of error rates for informed model selection.\footnote{\citet{tang-etal-2024-tofueval} provides a similar argument in favor of reporting error breakdown.} Similarly, for the metric developers, the breakdown highlights that TNR lags behind TPR by more than 10\% on 9 of 14 datasets; improving the ability of metrics to detect unattributable claims is a challenge. 

\subsection{\ais metrics incorrectly estimate the system error rate}
\label{sec:f3-poor-system-error-est}

Since the goal of AutoAIS evaluation metrics is to compare NLG systems, we study how accurate the automated metrics are in estimating the true (human-labeled) hallucination rate of the NLG systems.
For 12 datasets that contain generations from different systems, we group claims based on the system ($S$). For each system $S$, we report the \emph{bias} \citep{10.1145/3117807-review-on-quantification} of the AutoAIS metrics, which is the difference between the labeled error rate (percentage of claims labeled as unattributable) and the predicted error rate (percentage of claims predicted as ``unattributable'' by the \ais metric).
Additionally, on 10 of the 14 datasets where the claims are part of a longer response, we compute a response-level \emph{bias} as the difference between the response-level ground-truth error rate and the response-level predicted error rate.


\begin{figure}[t]
    \centering
    \includegraphics[width=0.48\textwidth]{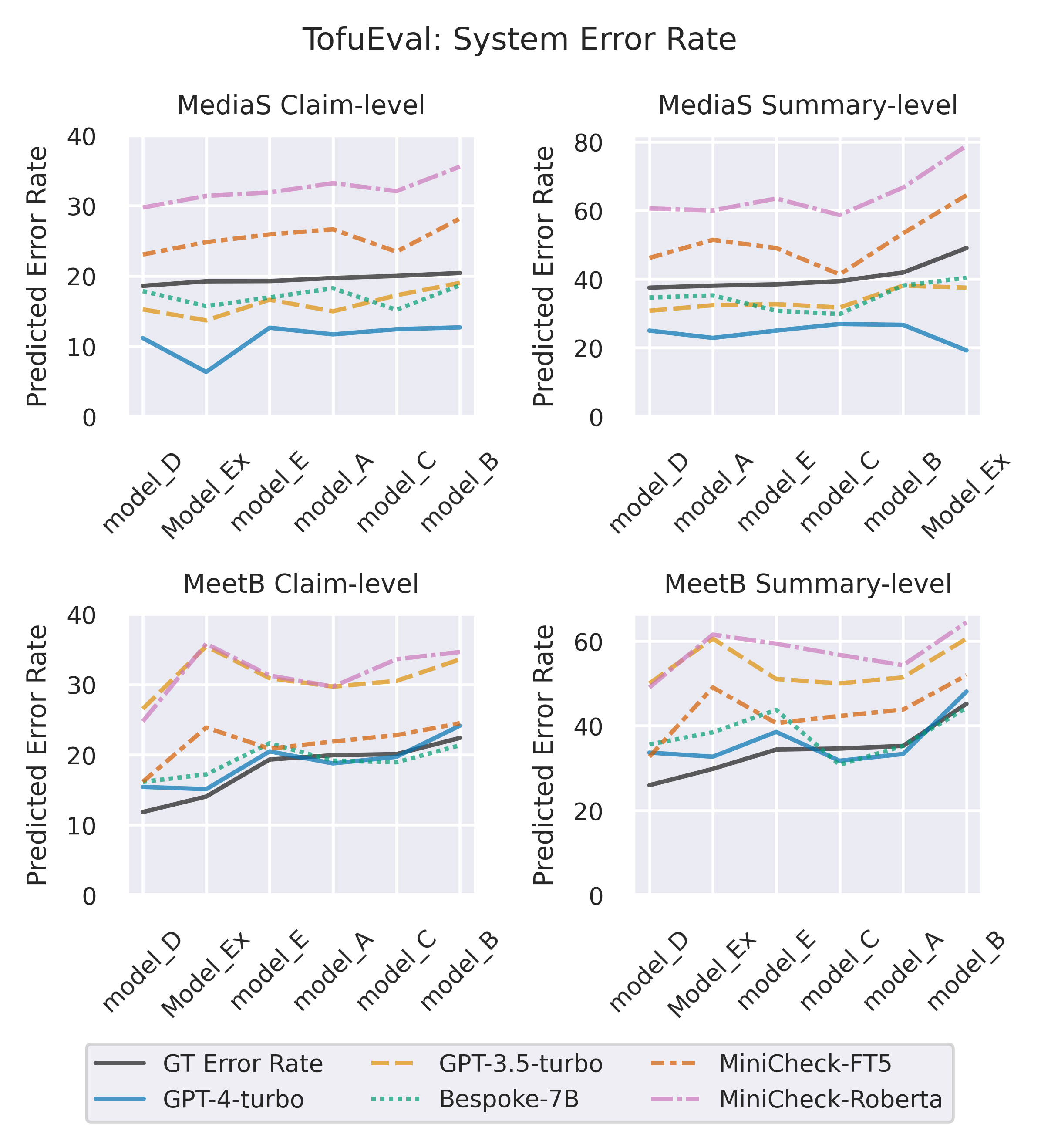}
    \caption{\textbf{Predicted system-level error rate on TofuEval.} Imperfect evaluators lead to differences in the ground truth and predicted error rate for different NLG systems. Claim-level misclassification leads to even greater quantification discrepancies in the summary-level attribution error rate.
    }
    \label{fig:tofueval-sys-err-rate}
    \vspace{-2ex}
\end{figure}

In Figure~\ref{fig:tofueval-sys-err-rate}, we highlight the bias of the metric on TofuEval-MediaSum and TofuEval-MeetingBank. From the claim-level error rates, we see that some metrics under-estimate the error rates of all the systems (\gptf, \gptt, and \bespoke) while others over-estimate the error rate (\mft and \mroberta). All 5 metrics have a balanced accuracy of 68-72\% on TofuEval-MediaS. Claim-level misclassification and inconsistencies compound when we compute response-level quantification error. On TofuEval-MediaS, the response-level biases (-29.8\% at worst for \gptf) are about twice the claim-level biases (-12.9\% at worst for \gptf). Similarly, in Figure~\ref{fig:ragtruth-sys-err-rate} (Appendix~\ref{app:ragtruth-quantification-error}), we see that the metrics consistently overestimate the system error rates on the RAGTruth dataset. Moreover, the magnitude of quantification error varies widely across 4 subsets of RAGTruth. We report the claim-level and response-level bias of the \ais metric on the 12 datasets in Appendix~\ref{app:quantification-error}.


Thus, the metrics sometimes overestimate and sometimes underestimate the error rate of the systems on different datasets. \textbf{This means that we can't know beforehand if a metric will assign reliable system-level scores on a new dataset.}



\subsection{Finding 4: Misleading conclusions about headroom}
\label{sec:f4-misleading-headroom}

Benchmarks are useful for development if there is room for improvement with future systems. If we want to replace human evaluation with automated metrics on new benchmarks, then the metrics must provide a reliable estimate of this "headroom". From Figure~\ref{fig:tofueval-sys-err-rate} and Table~\ref{tab:medias-system-instance-error}, we see that \gptf underestimates the headroom on TofuEval-MediaS by 12.3\% while \mroberta overestimates the headroom by 11.2\% despite both metrics having the same BAcc on the dataset. At the response level, this headroom estimation error grows in magnitude to -18.3\% for \gptf and +21.2\% for \mroberta.

Further, from Table~\ref{tab:ragtruth-system-instance-error}, we see that the headroom estimation is worse on smaller systems (7B params) than on larger systems (\gptt and gpt-4). For example, on RAGTruth-News, the \gptf evaluator misestimates headroom on the small systems by +7.3\% and on the large systems by +0.8\%. Thus, evaluators may unfairly score generations from smaller models leading to an inflated headroom.
\textbf{When creating a new benchmark, the evaluator must be validated to ensure that it correctly reflects the scope for improvement.}

\subsection{Finding 5: Misleading system rankings}
\label{sec:f5-poor-system-ranking}

The most important reason for using automated metrics is that they enable fast comparison of systems. A reliable metric ranks systems in the same order as the ranking determined by human labeling. Following \citet{mathur-etal-2020-tangled}, we identify which pairs of systems have statistically indistinguishable/distinguishable error rates. We then compare whether the \ais evaluator correctly predicts significant /insignificant differences between the system pairs. On 8 of 14 datasets (see Tables~\ref{tab:medias-instance-ranking-error},\ref{tab:meetb-instance-ranking-error},\ref{tab:expertqa-instance-ranking-error},\ref{tab:lfqa-instance-ranking-error},\ref{tab:ragtruth-instance-ranking-error}) with generations of at least 6 systems, \gptf orders 26\% system pairs incorrectly on average while \bespoke orders 20\% of pairs~\footnote{An ordering is incorrect if an indistinguishable system pair is predicted to have significant difference and vice versa. For the pairs where system A is significantly better than system B according to the annotated labels, none of the \ais metrics predict a ranking in the opposite direction (system B is significantly better than system A).} incorrectly on average. The limited number of examples per system in the dataset and the choice of systems with similar capabilities result in most of the system pairs being indistinguishable.

Further, unlike machine translation, datasets for factuality evaluation do not contain generations from a large number of systems; the datasets in \llmagg contain generations from at most 6 NLG systems while \citet{mathur-etal-2020-tangled} compare between 10-16 machine translation systems. This makes correlation analysis unreliable. For completeness, we report Kendall's $\tau$ and Pearson $\rho$ correlation coefficients in Appendix~\ref{app:ranking-error}. \textbf{Future AIS benchmarks should annotate more samples on the same input prompt (for statistically significant system differences) from more systems (for less noise in ranking correlation).}




\section{Analysis of metric biases}
\label{sec:metric-bias}

We identify two concerning biases that may affect evaluator predictions: (1) dependence on surface-level matches and (2) constraints due to context-window limitations. These biases may cause the evaluators to penalize desirable system outputs.

\subsection{Bias towards surface-level similarity}
\label{sec:b1-rouge-bias}

Evaluators heavily rely on surface-level matches between the claim and the document when making predictions.
We demonstrate this by studying the behavior of the \ais evaluators as the similarity between the claim and the document varies.

We measure similarity with ROUGE-2 precision~\citep{lin-2004-rouge}; this measures the fraction of claim bigrams that appear in the document. 
Following \citet{vu-etal-2024-foundational}, we partition the examples into 5 groups based on the task in the source dataset: (1) summarization tasks (`AggreFact-CNN', `AggreFact-XSum', `TofuEval-MediaS', `TofuEval-MeetB', `RAGTruth-CNN/DM', `RAGTruth-News'), (2) LLM response verification (`Reveal', `ClaimVerify', `FactCheck-GPT'), (3) Wikipedia verification (`Wice'), (4) Long-form QA (`ExpertQA', `Lfqa', `RAGTruth-MARCO'), and (5) Data2Text (`RAGTruth-Yelp'). Within each task group, we group examples into 5 bins based on percentiles of ROUGE-2 precision.

\begin{figure*}[htb]
    \centering
    \includegraphics[width=0.98\textwidth]{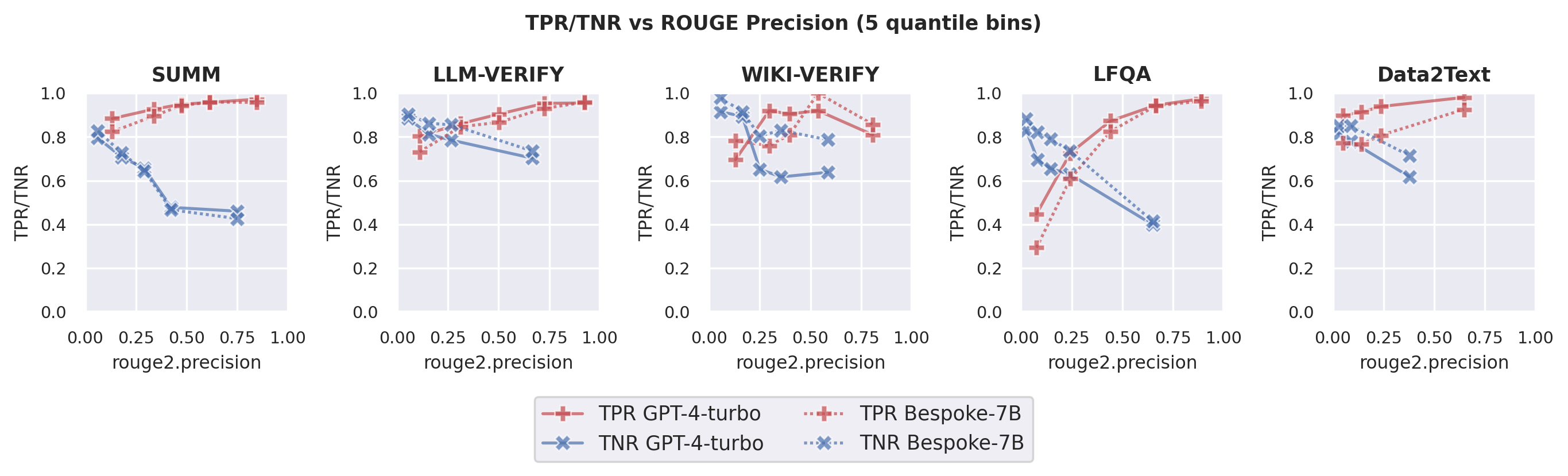}
    \vspace{-1ex}
    \caption{\textbf{TPR/TNR vs ROUGE-2 precision of AutoAIS evaluators:} ROUGE-2-precision is (anti-)correlated with true (negative)positive rate, i.e. metrics mislabel attributable generations with low ROUGE and unattributable generations with high ROUGE-2 precision.
    }
    \vspace{-2ex}
    \label{fig:rouge-calibration-task-group}
\end{figure*}

From Figure~\ref{fig:rouge-calibration-task-group}, we see that
the evaluators mislabel unattributable examples (have low TNR) on the high ROUGE examples. This trend is especially strong in the summarization and long-form QA groups, where the evaluators can detect unattributable claims with high ROUGE only half the time.
This highlights that evaluators may struggle to correctly identify small inconsistencies in otherwise heavily copied text.
Simultaneously, all evaluators have a trend of a low true positive rate on low ROUGE attributable claims; \ais evaluators penalize heavily paraphrased responses. This is a concern as the evaluators may penalize modern NLG systems for desirable behaviors such as avoiding verbatim copying and drawing valid conclusions. \textbf{Overall, the trends demonstrate that word overlap may be a significant component of the metric behavior.}

\subsection{Bias from context-size limitations}
\label{sec:b2-chunking-bias}

\ais evaluators with short context windows struggle when the claim connects different document parts. When using \ais evaluators, it is assumed that either (1) the metric has a sufficiently long context window to fit the document and the claim or (2) the metric chunks the document so as to fit it in the input length limit. As NLG systems improve at processing long documents and manipulating facts spread across a source document, it becomes more important for evaluation metrics to handle long evidence documents consistently.

To isolate the effect of chunking on \ais predictions, we compare the predictions of the \bespoke evaluator to the \bespoke evaluator with chunking enforced. 
From Table~\ref{tab:app-chunking-effect}, we see that in the subset of examples where chunking is applicable (document size > 500 words), \bespoke with chunked documents obtains a lower TPR and a higher TNR than the evaluator without chunking. This trade-off can be explained by the decrease in the fraction of examples predicted as "attributable"; \bespoke with chunked documents predicts the "attributable" label 6\% less frequently on average than the \bespoke evaluator with the full input.

To identify the examples where the evaluator predictions are most likely to be affected by chunking, we compute a score for every example that measures whether chunking reduces surface-level matches between the document chunk and the claim. In particular, borrowing notation from \S~\ref{sec:notation},
\begin{equation*}
    \setlength\abovedisplayskip{5pt}
    \setlength\belowdisplayskip{5pt}
    \begin{split}
        \texttt{R2-diff} = &\texttt{ROUGE-2}_{\texttt{prec}}(d, c) \\
                    & - \max_{k \in [1,K]} \texttt{ROUGE-2}_{\texttt{prec}}(d^{(k)}, c)
    \end{split}
\end{equation*}
where $\texttt{ROUGE-2}_{\texttt{prec}}$ is the fraction of claim bigrams that appear in the document. We expect examples with a nonzero value of $\texttt{R2-diff}$ reference words that do not all appear in one chunk, and thus, the claim is less likely to be verifiable on any single chunk. When the claim becomes unverifiable due to chunking, we expect the evaluator with chunked inputs to predict the label `unattributable' ($0$) more often than the evaluator with the full input. 

\begin{figure*}[htb]
    \centering
    \includegraphics[width=16cm]{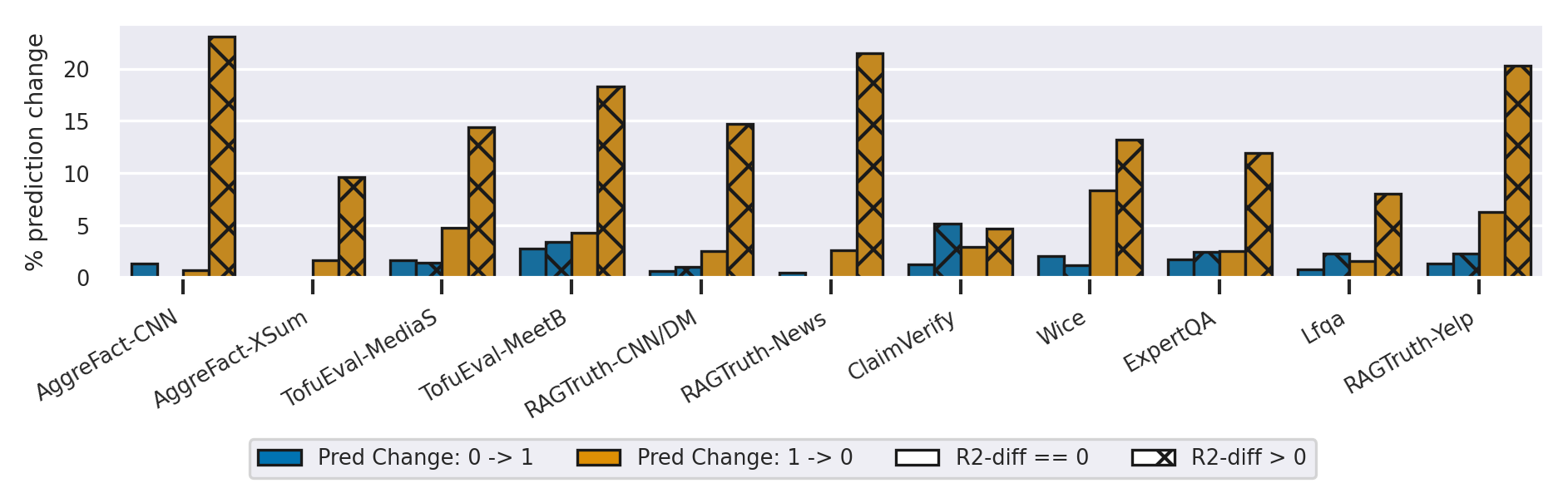}
    \vspace{-2em}
    \caption{\textbf{\texttt{R2-diff} vs rate of change in prediction with chunking:} The figure shows the change in predictions of the \bespoke evaluator to the same evaluator with documents chunked to 500 tokens. When chunking causes the overlap between the claim and the document to decrease (\texttt{R2-diff} > 0), the evaluator with chunking predicts the label `$0$' (unattributable) more frequently than the evaluator without chunking.}
    \label{fig:chunk-predictions-main}
\end{figure*}

In Figure~\ref{fig:chunk-predictions-main}, we plot how the original \bespoke metric predictions change when chunking is enforced.
We see that when $\texttt{R2-diff} > 0$, there is a marked increase in the predictions of the label `unattributable' ($0$). The rate is greater than 10\% on 8 of 11 datasets. The opposite change of prediction `$0$' with full context changing to label `$1$' with chunking is consistently less than 5\%. On the other hand, when $\texttt{R2-diff} == 0$, the rate of change in prediction is less than 10\% on 9 of 11 datasets. This could be attributed to noise in the metric predictions. \textbf{
Thus, evaluators that chunk their inputs are inherently disadvantaged when verifying attributable claims that reference distant parts of the input document.}
\section{Metric Adjustment}
\label{sec:adjustment}

\begin{table*}
\centering
\scriptsize
\begin{tabular}{llcccc}
\toprule
Model for Calibration & Source & No Adjustment & Adjusted Counts & Thres. tuning for zero bias & Thres. tuning for $\uparrow$BAcc \\
\midrule
\multirow{4}{*}{Cross-Validated} & CNN/DM & 3.9 (6.1) & 15.3 (21.5) & \cellcolor{green3}1.9 (4.0) & 14.8 (19.6) \\
& MARCO & 14.4 (22.3) & \cellcolor{green3}7.2 (12.1) & \cellcolor{green3}3.4 (6.7) & 20.2 (27.9) \\
& Recent News & 3.0 (5.9) & 10.5 (18.8) & \cellcolor{green3}2.3 (5.0) & 10.2 (18.8) \\
& Yelp & 15.4 (29.7) & 26.9 (43.7) & \cellcolor{green3}6.0 (13.1) & 19.3 (31.2) \\
\midrule
\multirow{4}{*}{\texttt{gpt-3.5-turbo-0613}} & CNN/DM & 4.5 (6.1) & 65.6 (86.5) & \cellcolor{green3}2.3 (4.7) & 37.0 (42.9) \\
& Recent News & 3.3 (6.0) & 11.2 (19.3) & \cellcolor{green3}1.7 (3.4) & 7.7 (15.4) \\
& MARCO & 16.0 (22.6) & 22.9 (32.6) & \cellcolor{green3}3.7 (7.2) & 30.0 (36.1) \\
& Yelp & 17.6 (31.2) & 52.8 (80.5) & \cellcolor{green3}6.7 (16.2) & 32.9 (46.7) \\
\bottomrule
\end{tabular}
\caption[\textbf{Comparison of adjustment methods on RAGTruth}]{\textbf{Comparison of adjustment methods on RAGTruth:} We report the bias in estimating the ground-truth system error (hallucination) rates using three adjustment methods.
In the upper section, we report cross-validated mean absolute bias by using one system for calibration and calculating the mean absolute bias over the remaining systems. Numbers in parentheses indicate the cross-validated worst-case bias. \colorbox{green3}{Green cells} indicate a decrease in bias relative to no adjustment. Tuning the evaluator threshold consistently reduces the bias in estimation over the held-out systems.
In the lower section, we report the mean absolute bias using the \gptt model for calibration (this is the model with the least ground-truth error rate).
See Tab~\ref{tab:app-ragtruth-metric-adjustment} for the full table.
}
\label{tab:ragtruth-metric-adjustment}
\end{table*}

As discussed in \S~\ref{sec:f3-poor-system-error-est}, the AutoAIS evaluators have a high bias in estimating the true error rate of NLG systems. We experiment with methods to reduce this bias and make the metrics more reliable in downstream applications. We assume a scenario where some human-labeled claim document pairs are available for calibration\footnote{This is a reasonable assumption when the metric is used to organize a new benchmark.}. For these experiments, we use the predictions and scores assigned by the \bespoke evaluator on examples from the RAGTruth datasets. We study three methods for reducing quantification bias: (1) post-hoc adjustment \citep{10.1145/1150402.1150423-adjust-count} that changes the predicted error rate based on the known TPR and FPR of the evaluator (details in Appendix~\ref{app:adjustment}), (2) threshold tuning to minimize the absolute bias and (3) threshold tuning to maximize BAcc.

In Table~\ref{tab:ragtruth-metric-adjustment}, we report the results of different methods for adjusting the predicted system error rate. We perform adjustment by using examples from one system for tuning the threshold / computing TPR and FPR and then computing the mean/worst absolute bias (magnitude) over all the remaining 5 systems. We report both the cross-validated mean and worst absolute bias. \textbf{We find that tuning to minimize the absolute bias consistently improves all four subsets of the RAGTruth dataset.} However, tuning to maximize BAcc leads to a degradation in both the mean and worst-case bias.

The "adjusted counts" approach is appealing to use if the \ais evaluator does not provide a scalar score and directly returns a label. The method shows an inconsistent reduction in the absolute bias. In particular, we find that using the best-performing system (\gptt for RAGTruth datasets) based on labeled error rate leads to poor estimation of the TPR and FPR of the evaluator. This is due to the low prevalence of the label `$0$' in examples of the best system. Simply adjusting counts based on these bad estimates leads to high bias on the remaining systems (see Table~\ref{tab:app-ragtruth-metric-adjustment} for the full table). When using the adjusted counts approach, we \textbf{advise against} using the system with the lowest error rate for calibration.

\section{Discussion}
\label{sec:discussion}

\subsection{For \ais Metric Developers}
\label{sec:d1-metric-dev}

Based on our findings in \S~\ref{sec:findings}, we urge developers of \ais evaluators to study and compare new approaches holistically. Our findings show that balanced accuracy can hide differences in the underlying behavior of different evaluators. (1) We advise that evaluator performance should be judged on the breakdown of true positive and negative rate (among evaluators with comparable balanced accuracy). AutoAIS metrics should be evaluated on the stability of TPR/TNR across datasets. (2) Quantification bias between predicted and ground-truth unattributable generation rate at the dataset and system levels should be reported. (3) Evaluators should report the rank correlation of NLG systems on the underlying dataset if available. These qualities establish how readily the evaluator can be applied to new domains and be used as a reliable stand-in for human annotations (though metric predictions should still be validated in new domains).

Since chunking long documents can make attributable claims unverifiable, when possible, emphasis should be placed on developing metrics that can process the entire evidence document without chunking. However, use cases such as \citet{wan2024positionalbiasfaithfulnesslongform} require judgment against long reference documents, and chunking becomes necessary. Thus, there is scope and reason to improve the ability of evaluators to correctly handle document chunking.


\subsection{For Benchmark Developers}
\label{sec:d2-benchmark-dev}

When benchmark curators use automated metrics for evaluation, it is necessary to validate the evaluators' performance against a human-annotated dataset. Based on the biases (\S~\ref{sec:metric-bias}) and findings (\S~\ref{sec:findings}), we encourage benchmark curators to:
\begin{enumerate}[nolistsep,noitemsep,leftmargin=16pt]
    \item Study evaluator behavior by strategically sampling examples from different buckets of the ROUGE precision distribution
    \item Validate the choice of using an evaluator that requires input document chunking by testing metric behavior on claims that require long-document reasoning. We highlight $\texttt{R2-diff}$ as an easy way to identify these claims.
    \item Validate the quantification bias of the evaluator on the human-annotated set. This allows for a better estimation of the actual headroom for improvement on the task.
    \item Validate the ranking and quantification bias on predictions from different NLG systems on the benchmark. Threshold tuning can be applied to reduce the bias at the system level.
\end{enumerate}

\subsection{For Hallucination Mitigation Research}
\label{sec:d3-mitigation-dev}

Based on our findings regarding error quantification bias at the system level, researchers working on hallucination mitigation should not use the absolute error rates predicted by \ais evaluators as the sole support for their research findings. Claims such as ``system A hallucinates less than system B" need to be paired with a validation of the evaluator predictions on claims from both systems. The quantification bias also highlights that automated evaluators alone are not an indicator of whether a dataset/task is solved/unsolved. Automated evaluators may under- or over-predict the system error rates.
These issues necessitate manual inspection of the evaluator's predictions to back claims based on automated metrics.

\section{Related Work}
\label{sec:relatedwork}

\noindent\textbf{Meta-Analysis of Automated Evaluation.}
\citet{nimah-etal-2023-nlg} suggest that NLG evaluator (fluency, coherence, consistency, relevance, etc) research should move beyond just measuring the correlation between human preferences and evaluator scores. They study the reliability of evaluators under domain shift and consistency with system rankings. Similar meta-analysis beyond correlation has been studied in extensively in machine translation \citep{mathur-etal-2020-tangled,kocmi-etal-2021-ship}. \citet{sai-etal-2021-perturbation} extend the checklist framework~\citep{ribeiro-etal-2020-beyond} to define consistency tests for NLG evaluators. In our work, we find that \ais evaluators are not yet reliable in certain downstream uses out-of-the-box and push for a holistic set of metrics for comparing evaluators.


\noindent\textbf{Meta-Analysis of \ais Evaluators.} Similar to \llmagg, AttributionBench~\citet{li-etal-2024-attributionbench} also aggregates datasets into an attribution evaluation benchmark. Error analysis by \citet{yue-etal-2023-automatic,li-etal-2024-attributionbench} also highlights the inability of \ais evaluators in judging nuanced claims. Cooroborating our findings about evaluator biases, concurrent work by \citet{ramprasad2024automaticfactualitymetricsmeasure} finds evidence that evaluators may be relying heavility on surface-level syntactic features. They find that evaluators can be ``gamed'' by making meaning-preserving edits to the claims.












\section*{Limitations}




Our analysis assumes that the datasets underlying \llmagg have highly accurate human annotations with little ambiguity. There is a potential confounder in our analysis that the human annotations may not be accurate or have significant room for ambiguity\citep{krishna-etal-2023-longeval,subbiah-etal-2024-storysumm,li-etal-2024-attributionbench}. In particular, \citet{li-etal-2024-attributionbench} highlight inbalances in the information accessible to humans vs \ais evaluators as a major source of error in evaluator predictions. We leave the reevaluation of this confounder for future work. We believe that a strong metric can be used in-the-loop to identify examples where the metric disagrees with the human label. These disagreements can help narrow down the set of examples with potentially ambiguous labels.

Our analysis is limited to the verification of claims against a single document. Complex claim verification might require multi-document verification \citep{chen-etal-2024-complex} which is currently out of the scope of this work.

Our analysis of system-level ranking is limited by the number of systems in the underlying dataset. In order to evaluate metrics on the consistency of system-level ranking, we need to collect responses from multiple, diverse NLG systems on a set of generation tasks and collect annotations of attributability. In prior work, the availability of predictions from multiple machine translation systems on a common evaluation set has allowed the machine translation community to study the reliability of automated metrics in ranking \citep{mathur-etal-2020-tangled}.

In our work, we identified that metrics make inconsistent misestimations on system-level factual accuracy. We do not propose any methods to fix these inconsistencies. A metric with perfect prediction accuracy will automatically solve the problem; however, the community needs a way to make reliable claims based on imperfect metrics in the interim.

\section*{Acknowledegments}
This work was supported in part by a gift from the USC-Amazon Center on Secure and Trusted Machine Learning, as well as
by the National Science
Foundation under Grant No. IIS-2403436.
Any opinions, findings, and conclusions or recommendations expressed
in this material are those of the author(s) and do not necessarily
reflect the views of the National Science Foundation. Feedback from Johnny Tian-Zheng Wei and Liyan Tang helped shape the project.

\bibliography{anthology,custom}

\appendix
\section{Appendix}
\label{sec:appendix}

{\small\listoftables}

\subsection{\llmagg Dataset Details}
\label{app:dataset-details}

Table~\ref{tab:app-dset-details} provides details of the 14 sub-datasets in \llmagg. Unlike \citet{tang-etal-2024-minicheck}, we keep the 4 subsets of RAGTruth \citep{niu-etal-2024-ragtruth} separate to highlight the underlying differences. The subsets are RAGTruth-CNN/DM, RAGTruth-Recent\_News (referred to as RAGTruth-News to save space), RAGTruth-MARCO, and RAGTruth-Yelp. We approximately follow \citet{vu-etal-2024-foundational} in the definition of the task groups. We mark the datasets where the claims are sourced from a longer response.

\begin{table*}
\centering
\scriptsize
\begin{tabular}{>{\raggedright}p{60pt} |c l c}
    \toprule
    Task & Dataset & Claim Source & Has Long Response? \\
    \midrule
    \multirow{6}{60pt}{Summarization} & AggreFact-CNN & \multirow{2}{*}{BART, T5, PEGASUS} & N \\
     & AggreFact-XSum & & N \\
     \cmidrule(lr){2-4}
     & TofuEval-MediaSum & \multirow{2}{*}{GPT-3.5-Turbo, Vicuna-7B, WizardLM-7B/13B/30B} & Y \\
     & TofuEval-MeetingBank & & Y \\
     \cmidrule(lr){2-4}
     & RAGTruth-CNN/DM & \multirow{2}{*}{GPT-3.5-turbo, GPT-4, Mistral-7b-Instruct, Llama-2-\{7B,13B,70B\}-chat} & Y \\
     & RAGTruth-Recent News & & Y \\
    \midrule
    \multirow{3}{60pt}{LLM Response Verification} & Reveal & Flan-PaLM-540B, text-davinci-003, Flan-UL2-20B & Y \\
     \cmidrule(lr){2-4}
     & ClaimVerify & Bing Chat, NeevaAI, perplexity.ai, YouChat & Y \\
     \cmidrule(lr){2-4}
     & FactCheckGPT & ChatGPT & N \\
     \midrule

     Wikipedia Verification & Wice & Human-written & N \\
     \midrule

     \multirow{3}{60pt}{Long-form QA} & ExpertQA & GPT4, Bing Chat & Y \\
     \cmidrule(lr){2-4}
     & LFQA & WebGPT, GPT-3.5, Alpaca-7b & Y \\
     \cmidrule(lr){2-4}
     & RAGTruth-MARCO & GPT-3.5-turbo, GPT-4, Mistral-7b-Instruct, Llama-2-\{7B,13B,70B\}-chat & Y \\
     \midrule

     Data2Text & RAGTruth-Yelp & GPT-3.5-turbo, GPT-4, Mistral-7b-Instruct, Llama-2-\{7B,13B,70B\}-chat & Y \\
     \bottomrule
\end{tabular}
\caption{\textbf{Description of the task types and claim sources in \llmagg}}
\label{tab:app-dset-details}
\end{table*}

\subsection{Recomputed Metric Performance}

Since we sub-divide RAGTruth into its component datasets, we report the recomputed balanced accuracy (BAcc) of the top \ais evaluators in Table~\ref{tab:full-bacc}. We report the breakdown by FPR in Table~\ref{tab:full-fpr} and FNR in Table~\ref{tab:full-fnr}.

\begin{table*}
\centering
\scriptsize
\setlength{\tabcolsep}{4pt}
\begin{tabular}{l|r|cccccccccccccc}
    \toprule
    Dataset & \multirow{2}{*}{Avg} & \multicolumn{2}{c}{\textsc{AggreFact}} & \multicolumn{2}{c}{\textsc{TofuEval}} & \multirow{2}{*}{\textsc{Wice}} & \multirow{2}{*}{\textsc{Reveal}} & \multirow{2}{*}{\shortstack{\textsc{Claim}\\\textsc{Verify}}} & \multirow{2}{*}{\shortstack{\textsc{Fact}\\\textsc{Check}}} & \multirow{2}{*}{\shortstack{\textsc{Expert}\\\textsc{QA}}} & \multirow{2}{*}{\textsc{Lfqa}} & \multicolumn{4}{c}{\textsc{RAGTruth}} \\
    \cmidrule(lr){3-4}\cmidrule(lr){5-6}\cmidrule(lr){13-16}
     & & \textsc{CNN} & \textsc{XSum} & \textsc{MediaS} & \textsc{MeetB} &  &  &  &  &  &  & \textsc{Marco} & \textsc{Yelp} & \textsc{CNN} & \textsc{News} \\
    \midrule
    gpt-4-turbo & 76.9 & 63.3 & 75.5 & 68.5 & 81.2 & 79.8 & 88.2 & 73.1 & 80.0 & 60.8 & 83.0 & 78.0 & 84.7 & 80.0 & 80.6 \\
    Bespoke-7B & 76.7 & 62.3 & 73.1 & 72.1 & 77.1 & 85.3 & 89.5 & 77.1 & 77.7 & 60.0 & 85.2 & 78.0 & 81.6 & 78.3 & 76.2 \\
    \phantom{a}+ chunk(500) & 75.9 & 64.5 & 72.6 & 72.0 & 75.8 & 77.3 & 89.5 & 77.1 & 77.7 & 59.8 & 85.0 & 77.9 & 78.7 & 78.4 & 76.1 \\
    MCheck-\textsc{Rbta} & 73.3 & 59.6 & 66.6 & 68.8 & 72.3 & 66.8 & 88.6 & 78.1 & 75.9 & 56.7 & 84.3 & 79.2 & 72.1 & 77.6 & 79.1 \\
    MCheck-\textsc{FT5} & 72.8 & 65.3 & 68.4 & 68.4 & 71.5 & 70.7 & 87.4 & 75.9 & 74.9 & 58.7 & 82.4 & 76.0 & 70.2 & 75.4 & 73.8 \\
    gpt-3.5-turbo & 72.2 & 64.8 & 71.0 & 66.3 & 74.8 & 70.5 & 85.1 & 72.1 & 74.6 & 58.3 & 77.8 & 70.2 & 77.4 & 70.8 & 76.7 \\
    AlignScore & 70.5 & 52.6 & 65.0 & 65.7 & 72.9 & 67.3 & 86.8 & 72.0 & 75.7 & 56.8 & 81.7 & 73.5 & 66.7 & 75.9 & 75.1 \\
    FactKB & 56.9 & 58.5 & 64.4 & 51.6 & 53.1 & 55.3 & 71.2 & 56.8 & 58.6 & 53.1 & 57.9 & 56.9 & 50.6 & 50.4 & 57.7 \\
    \bottomrule
\end{tabular}
\caption{\textbf{Balanced Accuracy of metrics on the dev set of \llmagg}}
\label{tab:full-bacc}
\end{table*}

\begin{table*}
\centering
\scriptsize
\setlength{\tabcolsep}{4pt}
\begin{tabular}{l|r|cccccccccccccc}
    \toprule
    Dataset & \multirow{2}{*}{Avg} & \multicolumn{2}{c}{\textsc{AggreFact}} & \multicolumn{2}{c}{\textsc{TofuEval}} & \multirow{2}{*}{\textsc{Wice}} & \multirow{2}{*}{\textsc{Reveal}} & \multirow{2}{*}{\shortstack{\textsc{Claim}\\\textsc{Verify}}} & \multirow{2}{*}{\shortstack{\textsc{Fact}\\\textsc{Check}}} & \multirow{2}{*}{\shortstack{\textsc{Expert}\\\textsc{QA}}} & \multirow{2}{*}{\textsc{Lfqa}} & \multicolumn{4}{c}{\textsc{RAGTruth}} \\
    \cmidrule(lr){3-4}\cmidrule(lr){5-6}\cmidrule(lr){13-16}
     & & \textsc{CNN} & \textsc{XSum} & \textsc{MediaS} & \textsc{MeetB} &  &  &  &  &  &  & \textsc{Marco} & \textsc{Yelp} & \textsc{CNN} & \textsc{News} \\
\midrule
GPT-4-turbo & 34.1 & 71.2 & 21.4 & 59.3 & 29.8 & 25.6 & 16.1 & 46.7 & 14.5 & 46.9 & 27.2 & 27.0 & 23.1 & 35.3 & 33.7 \\
Bespoke-7B & 30.6 & 69.5 & 25.3 & 47.5 & 36.3 & 13.7 & 15.0 & 36.0 & 9.3 & 33.2 & 21.4 & 25.2 & 17.6 & 36.9 & 42.2 \\
Bespoke-7B (cs=500) & 27.7 & 59.3 & 24.3 & 39.8 & 30.4 & 9.8 & 15.0 & 35.7 & 9.3 & 32.8 & 21.3 & 25.2 & 14.7 & 30.5 & 39.8 \\
MiniCheck-Roberta & 24.4 & 66.1 & 26.4 & 37.3 & 31.5 & 9.0 & 12.9 & 25.7 & 8.8 & 21.4 & 11.5 & 17.7 & 20.2 & 28.3 & 25.3 \\
MiniCheck-FT5 & 30.6 & 59.3 & 36.6 & 44.9 & 42.9 & 9.8 & 14.6 & 36.0 & 12.6 & 32.2 & 22.4 & 27.9 & 9.1 & 36.9 & 43.4 \\
GPT-3.5-turbo & 34.7 & 59.3 & 25.8 & 57.6 & 28.0 & 27.8 & 14.5 & 41.2 & 11.8 & 48.3 & 28.8 & 34.5 & 20.2 & 51.3 & 36.1 \\
AlignScore & 37.3 & 93.2 & 44.9 & 55.9 & 34.3 & 14.1 & 16.2 & 46.0 & 10.2 & 32.9 & 24.2 & 35.4 & 34.8 & 37.4 & 42.2 \\
FactKB & 64.8 & 78.0 & 17.0 & 91.2 & 80.6 & 59.8 & 15.6 & 78.3 & 32.3 & 74.5 & 77.7 & 44.2 & 90.8 & 96.3 & 71.1 \\
\bottomrule
\end{tabular}
\caption{\textbf{False positive rate (FPR) of metrics on the dev set of \llmagg}}
\label{tab:full-fpr}
\end{table*}

\begin{table*}
\centering
\scriptsize
\setlength{\tabcolsep}{4pt}
\begin{tabular}{l|r|cccccccccccccc}
    \toprule
    Dataset & \multirow{2}{*}{Avg} & \multicolumn{2}{c}{\textsc{AggreFact}} & \multicolumn{2}{c}{\textsc{TofuEval}} & \multirow{2}{*}{\textsc{Wice}} & \multirow{2}{*}{\textsc{Reveal}} & \multirow{2}{*}{\shortstack{\textsc{Claim}\\\textsc{Verify}}} & \multirow{2}{*}{\shortstack{\textsc{Fact}\\\textsc{Check}}} & \multirow{2}{*}{\shortstack{\textsc{Expert}\\\textsc{QA}}} & \multirow{2}{*}{\textsc{Lfqa}} & \multicolumn{4}{c}{\textsc{RAGTruth}} \\
    \cmidrule(lr){3-4}\cmidrule(lr){5-6}\cmidrule(lr){13-16}
     & & \textsc{CNN} & \textsc{XSum} & \textsc{MediaS} & \textsc{MeetB} &  &  &  &  &  &  & \textsc{Marco} & \textsc{Yelp} & \textsc{CNN} & \textsc{News} \\
\midrule
GPT-4-turbo & 12.1 & 2.2 & 27.7 & 3.7 & 7.8 & 14.8 & 7.5 & 7.1 & 25.6 & 31.5 & 6.9 & 16.9 & 7.4 & 4.7 & 5.1 \\
Bespoke-7B & 16.0 & 6.0 & 28.4 & 8.4 & 9.4 & 15.7 & 6.0 & 9.9 & 35.3 & 46.8 & 8.2 & 18.8 & 19.2 & 6.5 & 5.5 \\
Bespoke-7B (cs=500) & 20.5 & 11.8 & 30.5 & 16.2 & 18.0 & 35.7 & 6.0 & 10.2 & 35.3 & 47.7 & 8.7 & 18.9 & 27.9 & 12.7 & 8.0 \\
MiniCheck-Roberta & 29.0 & 14.8 & 40.4 & 25.1 & 23.8 & 57.4 & 9.9 & 18.0 & 39.5 & 65.1 & 19.9 & 23.9 & 35.6 & 16.5 & 16.4 \\
MiniCheck-FT5 & 23.8 & 10.0 & 26.6 & 18.3 & 14.1 & 48.7 & 10.6 & 12.2 & 37.6 & 50.3 & 12.7 & 20.1 & 50.5 & 12.2 & 8.9 \\
GPT-3.5-turbo & 21.0 & 11.0 & 32.2 & 9.7 & 22.5 & 31.3 & 15.3 & 14.6 & 39.1 & 35.1 & 15.7 & 25.1 & 25.1 & 7.1 & 10.6 \\
AlignScore & 21.6 & 1.5 & 25.1 & 12.7 & 20.0 & 51.3 & 10.1 & 10.1 & 38.3 & 53.5 & 12.3 & 17.6 & 31.8 & 10.8 & 7.6 \\
FactKB & 21.5 & 5.0 & 54.3 & 5.5 & 13.2 & 29.6 & 42.1 & 8.2 & 50.4 & 19.2 & 6.5 & 41.9 & 8.1 & 3.0 & 13.6 \\
\bottomrule
\end{tabular}
\caption{\textbf{False negative rate (FNR) of metrics on the dev set of \llmagg}}
\label{tab:full-fnr}
\end{table*}

\begin{table*}
\centering
\scriptsize
\setlength{\tabcolsep}{3pt}
\begin{tabular}{lllllllll}
\toprule
 &  & GPT-4-turbo & GPT-3.5-turbo & Bespoke-7B & Bespoke-7B (cs=500) & MiniCheck-FT5 & MiniCheck-Roberta & AlignScore \\
corr type & source &  &  &  &  &  &  &  \\
\midrule
\multirow[t]{7}{*}{Kendall's $\tau$} & ExpertQA & 0.73 & 0.60 & 0.47 & 0.60 & 0.60 & 0.87 & 0.73 \\
 & Lfqa & 0.87 & 0.87 & 0.87 & 0.87 & 0.87 & 0.87 & 0.87 \\
 & RAGTruth-CNN/DM & 1.00 & 1.00 & 0.87 & 0.73 & 0.73 & 0.47 & 0.73 \\
 & RAGTruth-News & 0.87 & 0.87 & 0.73 & 0.47 & 0.73 & 0.73 & 0.87 \\
 & RAGTruth-MARCO & 0.87 & 0.87 & 0.47 & 0.47 & 0.33 & 0.47 & 0.47 \\
 & RAGTruth-Yelp & 0.73 & 0.60 & 0.60 & 0.60 & 0.60 & 0.73 & 0.60 \\
 \cmidrule(lr){2-9}
 & Average & 0.84 & 0.80 & 0.67 & 0.62 & 0.64 & 0.69 & 0.71 \\
\midrule
\multirow[t]{7}{*}{Pearson $\rho$} & ExpertQA & 0.76 & 0.75 & 0.77 & 0.75 & 0.77 & 0.88 & 0.85 \\
 & Lfqa & 0.99 & 0.97 & 1.00 & 1.00 & 0.99 & 0.99 & 0.99 \\
 & RAGTruth-CNN/DM & 1.00 & 0.94 & 0.96 & 0.89 & 0.88 & 0.85 & 0.93 \\
 & RAGTruth-News & 0.93 & 0.92 & 0.91 & 0.93 & 0.81 & 0.73 & 0.94 \\
 & RAGTruth-MARCO & 0.90 & 0.92 & 0.83 & 0.84 & 0.80 & 0.78 & 0.83 \\
 & RAGTruth-Yelp & 0.98 & 0.92 & 0.91 & 0.92 & 0.78 & 0.87 & 0.85 \\
 \cmidrule(lr){2-9}
 & Average & 0.93 & 0.91 & 0.90 & 0.89 & 0.84 & 0.85 & 0.90 \\
\bottomrule
\end{tabular}
\caption[\textbf{System ranking correlation (claim-level labels)}]{\textbf{System ranking correlation (claim-level labels).} For 6 \llmagg datasets, we report the correlations between system rankings based on human-labeled error rate and predicted error rate by \ais evaluators. Each dataset has generations from 6 NLG systems. While the Pearson correlation coefficient is high the top evaluators, Kendall's $\tau$ is lower. The value of $\tau$ indicates that the evaluators make one-three ranking errors in each ranking of the 6 systems.}
\label{tab:ranking-corr}
\end{table*}

\begin{table*}
\centering
\scriptsize
\setlength{\tabcolsep}{3pt}
\begin{tabular}{lllllllll}
\toprule
 &  & GPT-4-turbo & GPT-3.5-turbo & Bespoke-7B & Bespoke-7B (cs=500) & MiniCheck-FT5 & MiniCheck-Roberta & AlignScore \\
corr type & source &  &  &  &  &  &  &  \\
\midrule
\multirow[t]{7}{*}{Kendall's $\tau$} & ExpertQA & 0.47 & 0.60 & 0.60 & 0.60 & 0.60 & 0.73 & 0.73 \\
 & Lfqa & 0.97 & 1.00 & 1.00 & 1.00 & 0.86 & 0.71 & 0.97 \\
 & RAGTruth-CNN/DM & 0.87 & 0.73 & 0.87 & 0.73 & 0.60 & 0.33 & 0.73 \\
 & RAGTruth-News & 1.00 & 1.00 & 1.00 & 0.87 & 0.87 & 0.73 & 1.00 \\
 & RAGTruth-MARCO & 1.00 & 0.73 & 0.73 & 0.73 & 0.60 & 0.73 & 0.73 \\
 & RAGTruth-Yelp & 0.60 & 0.47 & 0.47 & 0.33 & 0.47 & 0.47 & 0.33 \\
 \cmidrule(lr){2-9}
 & Average & 0.82 & 0.76 & 0.78 & 0.71 & 0.67 & 0.62 & 0.75 \\
\midrule
\multirow[t]{7}{*}{Pearson $\rho$} & ExpertQA & 0.71 & 0.80 & 0.74 & 0.69 & 0.76 & 0.82 & 0.89 \\
 & Lfqa & 0.99 & 0.97 & 0.98 & 0.97 & 0.97 & 0.92 & 0.99 \\
 & RAGTruth-CNN/DM & 1.00 & 0.93 & 0.96 & 0.89 & 0.89 & 0.81 & 0.93 \\
 & RAGTruth-News & 0.91 & 0.92 & 0.88 & 0.91 & 0.80 & 0.69 & 0.94 \\
 & RAGTruth-MARCO & 0.92 & 0.94 & 0.88 & 0.88 & 0.86 & 0.83 & 0.87 \\
 & RAGTruth-Yelp & 0.98 & 0.94 & 0.92 & 0.91 & 0.75 & 0.83 & 0.72 \\
 \cmidrule(lr){2-9}
 & Average & 0.92 & 0.92 & 0.89 & 0.88 & 0.84 & 0.82 & 0.89 \\
\bottomrule
\end{tabular}
\caption[\textbf{System ranking correlation (response-level labels)}]{\textbf{System ranking correlation (response-level labels).} For 6 \llmagg datasets, we report the correlations between system rankings based on human-labeled error rate and predicted error rate by \ais evaluators. The labels are aggregated at the response-level. Each dataset has generations from 6 NLG systems. While the Pearson correlation coefficient is high the top evaluators, Kendall's $\tau$ is lower indicating errors in system ranking.}
\label{tab:ranking-corr-summ}
\end{table*}

\subsection{Evaluator Ranking Performance}
\label{app:ranking-error}

On 6 of 14 datasets with generations from at least 6 systems and where the ground truth error rates aren't very close, we further measure the Kendall's $\tau$ rank correlation between the predicted ranking of systems by the evaluator and the human-labeled ranking (computed from the system error rates). From Table~\ref{tab:ranking-corr}, we see that the is at least one rank inversion in the ranking produced by the top metrics. \bespoke evaluator has up to 4 rank inversions (in ranking 6 systems) on two datasets. We see similar trends in rank correlation when labels are aggregated at the summary level (see Table~\ref{tab:ranking-corr-summ}).

However, in order to make the correlation coefficient useful, there is a need to build a benchmark with a larger number of systems with a wide range of ground truth error rates. The machine translation research community \citep{mathur-etal-2020-tangled} has built such resources by running annual shared tasks. Thus, for our main analysis, we count the number of ranking errors where insignificant ground truth difference between systems becomes significant with automated evaluators and vice versa.

\subsection{Evaluator Quantification Bias}
\label{app:quantification-error}

In Tables 8-29, we report the system-level predicted error rate and quantification bias (claim and response level), and system-level ranking errors for the \ais metrics on the 14 \llmagg datasets.

\begin{table*}
\centering
\scriptsize
\begin{tabular}{llllllll}
\toprule
 & label & GPT-4-turbo & GPT-3.5-turbo & Bespoke-7B & Bespoke-7B (cs=500) & MiniCheck-FT5 & MiniCheck-Rbta \\
dataset &  &  &  &  &  &  &  \\
\midrule
Wice & 67.0 (0.0) & 54.7 (-12.3) & 58.7 (-8.3) & 63.0 (-4.0) & 72.2 (5.2) & 76.5 (9.5) & 79.9 (12.9) \\
FactCheck-GPT & 82.7 (0.0) & 75.1 (-7.5) & 79.7 (-3.0) & 81.1 (-1.6) & 81.1 (-1.6) & 78.8 (-3.9) & 82.2 (-0.5) \\
\bottomrule
\end{tabular}
\caption{\textbf{Wice and FactCheck:} Quantification bias of metrics}
\label{tab:wice-factcheck-system-instance-error}
\end{table*}

\begin{table*}
\centering
\tiny
\begin{tabular}{llllllll}
\toprule
 & label & GPT-4-turbo & GPT-3.5-turbo & Bespoke-7B & Bespoke-7B (cs=500) & MiniCheck-FT5 & MiniCheck-Rbta \\
System Name &  &  &  &  &  &  &  \\
\midrule
BART & 17.9 (0.0) & 6.4 (-11.5) & 14.5 (-3.4) & 9.8 (-8.1) & 17.5 (-0.4) & 15.8 (-2.1) & 18.8 (0.9) \\
Pegasus & 9.6 (0.0) & 4.0 (-5.6) & 16.0 (6.4) & 8.8 (-0.8) & 16.0 (6.4) & 12.8 (3.2) & 20.8 (11.2) \\
PegasusDynamic & 6.0 (0.0) & 4.0 (-2.0) & 20.0 (14.0) & 6.0 (0.0) & 10.0 (4.0) & 8.0 (2.0) & 6.0 (0.0) \\
T5 & 4.0 (0.0) & 8.0 (4.0) & 8.0 (4.0) & 10.0 (6.0) & 10.0 (6.0) & 14.0 (10.0) & 12.0 (8.0) \\
\midrule
Headroom & 4.0 (0.0) & 4.0 (0.0) & 8.0 (4.0) & 6.0 (2.0) & 10.0 (6.0) & 8.0 (4.0) & 6.0 (2.0) \\
\bottomrule
\end{tabular}
\caption{\textbf{AggreFact-CNN: Predicted instance-level error rates for systems.} Quantification bias in paratheses.}
\label{tab:aggrefact-cnn-system-instance-error}
\end{table*}


\begin{table*}
\centering
\tiny
\begin{tabular}{llllllll}
\toprule
 & label & GPT-4-turbo & GPT-3.5-turbo & Bespoke-7B & Bespoke-7B (cs=500) & MiniCheck-FT5 & MiniCheck-Roberta \\
System Name &  &  &  &  &  &  &  \\
\midrule
BART & 49.0 (0.0) & 53.3 (4.3) & 53.1 (4.1) & 52.8 (3.8) & 54.4 (5.4) & 45.4 (-3.6) & 57.7 (8.7) \\
Pegasus & 52.0 (0.0) & 48.0 (-4.0) & 50.7 (-1.3) & 36.0 (-16.0) & 37.3 (-14.7) & 38.7 (-13.3) & 48.0 (-4.0) \\
\midrule
Headroom & 49.0 (0.0) & 48.0 (-1.0) & 50.7 (1.7) & 36.0 (-13.0) & 37.3 (-11.7) & 38.7 (-10.3) & 48.0 (-1.0) \\
\bottomrule
\end{tabular}
\caption{\textbf{AggreFact-XSum: Predicted instance-level error rates for systems.} Quantification bias in paratheses.}
\label{tab:aggrefact-xsum-system-instance-error}
\end{table*}


\begin{table*}
\centering
\tiny
\begin{tabular}{llllllll}
\toprule
 & GT Label & GPT-4-turbo & GPT-3.5-turbo & Bespoke-7B & Bespoke-7B (cs=500) & MiniCheck-FT5 & MiniCheck-Roberta \\
System Name &  &  &  &  &  &  &  \\
\midrule
Model-Extra & 19.2 (0.0) & 6.3 (-12.9) & 13.7 (-5.6) & 15.7 (-3.5) & 21.3 (2.0) & 24.8 (5.6) & 31.4 (12.2) \\
model\_A & 19.7 (0.0) & 11.7 (-8.0) & 15.0 (-4.7) & 18.2 (-1.5) & 24.1 (4.4) & 26.6 (6.9) & 33.2 (13.5) \\
model\_B & 20.4 (0.0) & 12.7 (-7.7) & 19.0 (-1.4) & 18.7 (-1.8) & 29.9 (9.5) & 28.2 (7.7) & 35.6 (15.1) \\
model\_C & 20.0 (0.0) & 12.4 (-7.6) & 17.2 (-2.8) & 15.2 (-4.8) & 24.5 (4.5) & 23.4 (3.4) & 32.1 (12.1) \\
model\_D & 18.6 (0.0) & 11.2 (-7.4) & 15.2 (-3.3) & 17.8 (-0.7) & 24.9 (6.3) & 23.0 (4.5) & 29.7 (11.2) \\
model\_E & 19.3 (0.0) & 12.6 (-6.6) & 16.6 (-2.7) & 16.9 (-2.3) & 24.6 (5.3) & 25.9 (6.6) & 31.9 (12.6) \\
\midrule
Headroom & 18.6 (0.0) & 6.3 (-12.3) & 13.7 (-4.9) & 15.2 (-3.4) & 21.3 (2.7) & 23.0 (4.5) & 29.7 (11.2) \\
\bottomrule
\end{tabular}
\caption{\textbf{TofuEval-MediaSum: Predicted claim-level error rates for systems.} Quantification bias in paratheses.}
\label{tab:medias-system-instance-error}
\end{table*}

\begin{table*}
\centering
\tiny
\begin{tabular}{l|ccc|ccc|ccc|ccc|ccc|ccc|ccc|}
\toprule
GT Order & \multicolumn{3}{c|}{GPT-4-turbo} & \multicolumn{3}{c|}{GPT-3.5-turbo} & \multicolumn{3}{c|}{Bespoke-7B} & \multicolumn{3}{c|}{Bespoke-7B (cs=500)} & \multicolumn{3}{c|}{MiniCheck-FT5} & \multicolumn{3}{c|}{MiniCheck-Roberta} & \multicolumn{3}{c|}{AlignScore} \\
 & > & = & < & > & = & < & > & = & < & > & = & < & > & = & < & > & = & < & > & = & < \\
\midrule
= & 1 & 10 & 4 & 0 & 15 & 0 & 0 & 15 & 0 & 0 & 14 & 1 & 0 & 15 & 0 & 0 & 15 & 0 & 0 & 15 & 0 \\
< & 0 & 0 & 0 & 0 & 0 & 0 & 0 & 0 & 0 & 0 & 0 & 0 & 0 & 0 & 0 & 0 & 0 & 0 & 0 & 0 & 0 \\
\midrule
\%Err & \multicolumn{3}{c|}{33.3} & \multicolumn{3}{c|}{0.0} & \multicolumn{3}{c|}{0.0} & \multicolumn{3}{c|}{6.7} & \multicolumn{3}{c|}{0.0} & \multicolumn{3}{c|}{0.0} & \multicolumn{3}{c|}{0.0} \\
\%Maj. Err & \multicolumn{3}{c|}{0.0} & \multicolumn{3}{c|}{0.0} & \multicolumn{3}{c|}{0.0} & \multicolumn{3}{c|}{0.0} & \multicolumn{3}{c|}{0.0} & \multicolumn{3}{c|}{0.0} & \multicolumn{3}{c|}{0.0} \\
\bottomrule
\end{tabular}
\caption[\textbf{TofuEval-MediaSum: Inconsistency in system-pair ranking based on claim-level error rates for systems}]{\textbf{TofuEval-MediaSum: Inconsistency in system-pair ranking based on claim-level error rates for systems.} We report a confusion matrix of pairwise system ranking decisions. We measure inconsistencies between the ranking based on the labeled error rate and the ranking based on the predicted error rate. For a system pair (s1, s2),  `=' indicates no significant difference between s1 and s2, `<' indicates s1 has a lower error rate than s2, and `>' indicates s1 has a higher error rate than s2. When a metric predicts a significant but opposite ranking between a pair, we count it as a
Major Error. Significance is computed with the two-proportion z-test and p\_value < 0.05.}
\label{tab:medias-instance-ranking-error}
\end{table*}

\begin{table*}
\centering
\tiny
\begin{tabular}{llllllll}
\toprule
 & label & GPT-4-turbo & GPT-3.5-turbo & Bespoke-7B & Bespoke-7B (cs=500) & MiniCheck-FT5 & MiniCheck-Roberta \\
System Name &  &  &  &  &  &  &  \\
\midrule
Model-Extra & 49.0 (0.0) & 19.2 (-29.8) & 37.5 (-11.5) & 40.4 (-8.7) & 55.8 (6.7) & 64.4 (15.4) & 78.8 (29.8) \\
model\_A & 38.1 (0.0) & 22.9 (-15.2) & 32.4 (-5.7) & 35.2 (-2.9) & 44.8 (6.7) & 51.4 (13.3) & 60.0 (21.9) \\
model\_B & 41.9 (0.0) & 26.7 (-15.2) & 38.1 (-3.8) & 38.1 (-3.8) & 56.2 (14.3) & 53.3 (11.4) & 66.7 (24.8) \\
model\_C & 39.4 (0.0) & 26.9 (-12.5) & 31.7 (-7.7) & 29.8 (-9.6) & 48.1 (8.7) & 41.3 (1.9) & 58.7 (19.2) \\
model\_D & 37.5 (0.0) & 25.0 (-12.5) & 30.8 (-6.7) & 34.6 (-2.9) & 45.2 (7.7) & 46.2 (8.7) & 60.6 (23.1) \\
model\_E & 38.5 (0.0) & 25.0 (-13.5) & 32.7 (-5.8) & 30.8 (-7.7) & 46.2 (7.7) & 49.0 (10.6) & 63.5 (25.0) \\
\midrule
Headroom & 37.5 (0.0) & 19.2 (-18.3) & 30.8 (-6.7) & 29.8 (-7.7) & 44.8 (7.3) & 41.3 (3.8) & 58.7 (21.2) \\
\bottomrule
\end{tabular}
\caption{\textbf{TofuEval-MediaSum: Predicted summary-level error rates for systems.} Quantification bias in paratheses.}
\label{tab:medias-system-summary-error}
\end{table*}

\begin{table*}
\centering
\tiny
\begin{tabular}{llllllll}
\toprule
 & GT Error Rate & GPT-4-turbo & GPT-3.5-turbo & Bespoke-7B & Bespoke-7B (cs=500) & MiniCheck-FT5 & MiniCheck-Roberta \\
System &  &  &  &  &  &  &  \\
\midrule
Model-Extra & 14.0 (0.0) & 15.1 (1.1) & 35.4 (21.4) & 17.2 (3.2) & 29.1 (15.1) & 23.9 (9.8) & 35.8 (21.8) \\
model-A & 19.9 (0.0) & 18.8 (-1.2) & 29.7 (9.8) & 19.1 (-0.8) & 28.5 (8.6) & 21.9 (2.0) & 29.7 (9.8) \\
model-B & 22.4 (0.0) & 24.1 (1.7) & 33.6 (11.2) & 21.3 (-1.0) & 27.6 (5.2) & 24.5 (2.1) & 34.6 (12.2) \\
model-C & 20.1 (0.0) & 19.7 (-0.4) & 30.5 (10.4) & 18.9 (-1.2) & 27.4 (7.3) & 22.8 (2.7) & 33.6 (13.5) \\
model-D & 11.8 (0.0) & 15.4 (3.6) & 26.5 (14.7) & 16.1 (4.3) & 22.2 (10.4) & 16.1 (4.3) & 24.7 (12.9) \\
model-E & 19.3 (0.0) & 20.5 (1.2) & 30.9 (11.6) & 21.6 (2.3) & 27.4 (8.1) & 20.8 (1.5) & 31.3 (12.0) \\
\midrule
Headroom & 11.8 (0.0) & 15.1 (3.3) & 26.5 (14.7) & 16.1 (4.3) & 22.2 (10.4) & 16.1 (4.3) & 24.7 (12.9) \\
\bottomrule
\end{tabular}
\caption{\textbf{TofuEval-MeetingBank: Predicted claim-level error rates for systems.} Quantification bias in paratheses.}
\label{tab:meetb-system-instance-error}
\end{table*}

\begin{table*}
\centering
\tiny
\begin{tabular}{l|ccc|ccc|ccc|ccc|ccc|ccc|ccc|}
\toprule
GT Order & \multicolumn{3}{c|}{GPT-4-turbo} & \multicolumn{3}{c|}{GPT-3.5-turbo} & \multicolumn{3}{c|}{Bespoke-7B} & \multicolumn{3}{c|}{Bespoke-7B (cs=500)} & \multicolumn{3}{c|}{MiniCheck-FT5} & \multicolumn{3}{c|}{MiniCheck-Roberta} & \multicolumn{3}{c|}{AlignScore} \\
 & > & = & < & > & = & < & > & = & < & > & = & < & > & = & < & > & = & < & > & = & < \\
\midrule
= & 0 & 10 & 0 & 0 & 9 & 1 & 0 & 10 & 0 & 0 & 10 & 0 & 0 & 9 & 1 & 0 & 9 & 1 & 1 & 8 & 1 \\
< & 0 & 3 & 2 & 0 & 5 & 0 & 0 & 5 & 0 & 0 & 5 & 0 & 0 & 4 & 1 & 0 & 3 & 2 & 0 & 2 & 3 \\
\midrule
\%Err & \multicolumn{3}{c|}{20.0} & \multicolumn{3}{c|}{40.0} & \multicolumn{3}{c|}{33.3} & \multicolumn{3}{c|}{33.3} & \multicolumn{3}{c|}{33.3} & \multicolumn{3}{c|}{26.7} & \multicolumn{3}{c|}{26.7} \\
\%Maj. Err & \multicolumn{3}{c|}{0.0} & \multicolumn{3}{c|}{0.0} & \multicolumn{3}{c|}{0.0} & \multicolumn{3}{c|}{0.0} & \multicolumn{3}{c|}{0.0} & \multicolumn{3}{c|}{0.0} & \multicolumn{3}{c|}{0.0} \\
\bottomrule
\end{tabular}
\caption[\textbf{TofuEval-MeetingBank: Inconsistency in system-pair ranking based on claim-level error rates for systems}]{\textbf{TofuEval-MeetingBank: Inconsistency in system-pair ranking based on claim-level error rates for systems.} We report a confusion matrix of pairwise system ranking decisions. We measure inconsistencies between the ranking based on the labeled error rate and the ranking based on the predicted error rate. For a system pair (s1, s2),  `=' indicates no significant difference between s1 and s2, `<' indicates s1 has a lower error rate than s2, and `>' indicates s1 has a higher error rate than s2. When a metric predicts a significant but opposite ranking between a pair, we count it as a
Major Error. Significance is computed with the two-proportion z-test and p\_value < 0.05.}
\label{tab:meetb-instance-ranking-error}
\end{table*}

\begin{table*}
\centering
\tiny
\begin{tabular}{llllllll}
\toprule
 & label & GPT-4-turbo & GPT-3.5-turbo & Bespoke-7B & Bespoke-7B (cs=500) & MiniCheck-FT5 & MiniCheck-Roberta \\
System Name &  &  &  &  &  &  &  \\
\midrule
Model-Extra & 29.8 (0.0) & 32.7 (2.9) & 60.6 (30.8) & 38.5 (8.7) & 57.7 (27.9) & 49.0 (19.2) & 61.5 (31.7) \\
model\_A & 35.2 (0.0) & 33.3 (-1.9) & 51.4 (16.2) & 35.2 (0.0) & 52.4 (17.1) & 43.8 (8.6) & 54.3 (19.0) \\
model\_B & 45.2 (0.0) & 48.1 (2.9) & 60.6 (15.4) & 44.2 (-1.0) & 53.8 (8.7) & 51.9 (6.7) & 64.4 (19.2) \\
model\_C & 34.6 (0.0) & 31.7 (-2.9) & 50.0 (15.4) & 30.8 (-3.8) & 46.2 (11.5) & 42.3 (7.7) & 56.7 (22.1) \\
model\_D & 26.0 (0.0) & 33.7 (7.7) & 50.0 (24.0) & 35.6 (9.6) & 48.1 (22.1) & 32.7 (6.7) & 49.0 (23.1) \\
model\_E & 34.4 (0.0) & 38.5 (4.2) & 51.0 (16.7) & 43.8 (9.4) & 53.1 (18.8) & 40.6 (6.2) & 59.4 (25.0) \\
\midrule
Headroom & 26.0 (0.0) & 31.7 (5.8) & 50.0 (24.0) & 30.8 (4.8) & 46.2 (20.2) & 32.7 (6.7) & 49.0 (23.1) \\
\bottomrule
\end{tabular}
\caption{\textbf{TofuEval-MeetingBank: Predicted summary-level error rates for systems.} Quantification bias in paratheses.}
\label{tab:meetb-system-summary-error}
\end{table*}

\begin{table*}
\centering
\tiny
\begin{tabular}{llllllll}
\toprule
 & label & GPT-4-turbo & GPT-3.5-turbo & Bespoke-7B & Bespoke-7B (cs=500) & MiniCheck-FT5 & MiniCheck-Roberta \\
System &  &  &  &  &  &  &  \\
\midrule
Flan-PaLM-540B & 67.4 (0.0) & 59.7 (-7.7) & 61.0 (-6.3) & 58.5 (-8.8) & 58.5 (-8.8) & 61.2 (-6.1) & 57.8 (-9.6) \\
Flan-UL2-20B & 79.4 (0.0) & 68.8 (-10.6) & 71.4 (-8.0) & 68.7 (-10.7) & 68.7 (-10.7) & 67.8 (-11.6) & 70.0 (-9.4) \\
GPT-3 & 76.2 (0.0) & 63.0 (-13.2) & 68.6 (-7.6) & 65.6 (-10.6) & 65.6 (-10.6) & 68.3 (-7.9) & 72.5 (-3.7) \\
\midrule
Headroom & 67.4 (0.0) & 59.7 (-7.7) & 61.0 (-6.3) & 58.5 (-8.8) & 58.5 (-8.8) & 61.2 (-6.1) & 57.8 (-9.6) \\
\bottomrule
\end{tabular}
\caption{\textbf{Reveal: Predicted instance-level error rates for systems.} Quantification bias in paratheses.}
\label{tab:reveal-system-instance-error}
\end{table*}

\begin{table*}
\centering
\tiny
\begin{tabular}{llllllll}
\toprule
 & label & GPT-4-turbo & GPT-3.5-turbo & Bespoke-7B & Bespoke-7B (cs=500) & MiniCheck-FT5 & MiniCheck-Roberta \\
System Name &  &  &  &  &  &  &  \\
\midrule
Flan-PaLM-540B & 74.7 (0.0) & 72.1 (-2.6) & 74.7 (0.0) & 69.5 (-5.2) & 69.5 (-5.2) & 69.5 (-5.2) & 69.5 (-5.2) \\
Flan-UL2-20B & 84.2 (0.0) & 76.6 (-7.6) & 79.5 (-4.7) & 76.6 (-7.6) & 76.6 (-7.6) & 77.2 (-7.0) & 80.1 (-4.1) \\
GPT-3 & 78.9 (0.0) & 66.3 (-12.6) & 74.2 (-4.7) & 68.4 (-10.5) & 68.4 (-10.5) & 72.1 (-6.8) & 77.4 (-1.6) \\
\midrule
Headroom & 74.7 (0.0) & 66.3 (-8.4) & 74.2 (-0.5) & 68.4 (-6.3) & 68.4 (-6.3) & 69.5 (-5.2) & 69.5 (-5.2) \\
\bottomrule
\end{tabular}
\caption{\textbf{Reveal: Predicted summary-level error rates for systems.} Quantification bias in paratheses.}
\label{tab:reveal-system-summary-error}
\end{table*}

\begin{table*}
\centering
\tiny
\begin{tabular}{llllllll}
\toprule
 & label & GPT-4-turbo & GPT-3.5-turbo & Bespoke-7B & Bespoke-7B (cs=500) & MiniCheck-FT5 & MiniCheck-Roberta \\
System &  &  &  &  &  &  &  \\
\midrule
bing\_chat & 9.4 & 6.1 (-3.3) & 10.7 (1.2) & 10.7 (1.2) & 9.8 (0.4) & 11.5 (2.0) & 13.1 (3.7) \\
neeva & 27.3 & 23.4 (-3.9) & 31.2 (3.9) & 26.3 (-1.0) & 28.6 (1.3) & 29.3 (2.0) & 34.5 (7.2) \\
perplexity & 30.7 & 19.6 (-11.2) & 29.4 (-1.4) & 26.6 (-4.1) & 26.8 (-3.9) & 28.0 (-2.7) & 37.2 (6.5) \\
you & 31.3 & 34.3 (3.0) & 32.8 (1.5) & 28.4 (-3.0) & 25.4 (-6.0) & 29.9 (-1.5) & 47.8 (16.4) \\
\midrule
Headroom & 9.4 (0.0) & 6.1 (-3.3) & 10.7 (1.2) & 10.7 (1.2) & 9.8 (0.4) & 11.5 (2.0) & 13.1 (3.7) \\
\bottomrule
\end{tabular}
\caption{\textbf{ClaimVerify: Predicted instance-level error rates for systems.} Quantification bias in paratheses.}
\label{tab:claimverify-system-instance-error}
\end{table*}

\begin{table*}
\centering
\tiny
\begin{tabular}{llllllll}
\toprule
 & label & GPT-4-turbo & GPT-3.5-turbo & Bespoke-7B & Bespoke-7B (cs=500) & MiniCheck-FT5 & MiniCheck-Roberta \\
System Name &  &  &  &  &  &  &  \\
\midrule
bing\_chat & 16.3 (0.0) & 11.4 (-4.9) & 18.7 (2.4) & 18.7 (2.4) & 16.3 (0.0) & 20.3 (4.1) & 19.5 (3.3) \\
neeva & 51.9 (0.0) & 45.3 (-6.6) & 56.6 (4.7) & 53.8 (1.9) & 56.6 (4.7) & 59.4 (7.5) & 61.3 (9.4) \\
perplexity & 53.6 (0.0) & 38.6 (-15.0) & 55.7 (2.1) & 52.9 (-0.7) & 50.7 (-2.9) & 54.3 (0.7) & 64.3 (10.7) \\
you & 38.6 (0.0) & 45.5 (6.8) & 40.9 (2.3) & 38.6 (0.0) & 36.4 (-2.3) & 40.9 (2.3) & 61.4 (22.7) \\
\midrule
Headroom & 16.3 (0.0) & 11.4 (-4.9) & 18.7 (2.4) & 18.7 (2.4) & 16.3 (0.0) & 20.3 (4.1) & 19.5 (3.3) \\
\bottomrule
\end{tabular}
\caption{\textbf{ClaimVerify: Predicted summary-level error rates for systems.} Quantification bias in paratheses.}
\label{tab:claimverify-system-summary-error}
\end{table*}

\begin{table*}
\centering
\tiny
\begin{tabular}{llllllll}
\toprule
 & label & GPT-4-turbo & GPT-3.5-turbo & Bespoke-7B & Bespoke-7B (cs=500) & MiniCheck-FT5 & MiniCheck-Roberta \\
System Name &  &  &  &  &  &  &  \\
\midrule
bing\_chat & 16.7 (0.0) & 34.1 (17.4) & 40.2 (23.5) & 44.0 (27.3) & 49.4 (32.7) & 49.7 (33.0) & 57.1 (40.4) \\
gpt4 & 27.4 (0.0) & 62.1 (34.7) & 54.7 (27.4) & 73.7 (46.3) & 75.8 (48.4) & 78.9 (51.6) & 90.5 (63.2) \\
post\_hoc\_gs\_gpt4 & 22.1 (0.0) & 52.8 (30.7) & 54.1 (32.0) & 74.8 (52.7) & 74.8 (52.7) & 73.8 (51.7) & 86.3 (64.2) \\
post\_hoc\_sphere\_gpt4 & 33.5 (0.0) & 53.8 (20.4) & 53.8 (20.4) & 72.8 (39.3) & 72.8 (39.3) & 71.7 (38.3) & 92.6 (59.2) \\
rr\_gs\_gpt4 & 11.7 (0.0) & 8.7 (-3.0) & 11.8 (0.1) & 16.7 (5.1) & 16.8 (5.2) & 23.3 (11.7) & 31.7 (20.1) \\
rr\_sphere\_gpt4 & 20.3 (0.0) & 9.8 (-10.4) & 17.1 (-3.1) & 18.7 (-1.6) & 18.9 (-1.4) & 28.5 (8.3) & 46.9 (26.6) \\
\midrule
Headroom & 11.7 (0.0) & 8.7 (-3.0) & 11.8 (0.1) & 16.7 (5.1) & 16.8 (5.2) & 23.3 (11.7) & 31.7 (20.1) \\
\bottomrule
\end{tabular}
\caption{\textbf{ExpertQA: Predicted claim-level error rates for systems.} Quantification bias in paratheses.}
\label{tab:expertqa-system-instance-error}
\end{table*}

\begin{table*}
\centering
\tiny
\begin{tabular}{l|ccc|ccc|ccc|ccc|ccc|ccc|ccc|}
\toprule
GT Order & \multicolumn{3}{c|}{GPT-4-turbo} & \multicolumn{3}{c|}{GPT-3.5-turbo} & \multicolumn{3}{c|}{Bespoke-7B} & \multicolumn{3}{c|}{Bespoke-7B (cs=500)} & \multicolumn{3}{c|}{MiniCheck-FT5} & \multicolumn{3}{c|}{MiniCheck-Roberta} & \multicolumn{3}{c|}{AlignScore} \\
 & > & = & < & > & = & < & > & = & < & > & = & < & > & = & < & > & = & < & > & = & < \\
\midrule
= & 1 & 2 & 2 & 1 & 2 & 2 & 1 & 2 & 2 & 1 & 2 & 2 & 1 & 2 & 2 & 1 & 2 & 2 & 0 & 3 & 2 \\
< & 0 & 2 & 8 & 0 & 1 & 9 & 0 & 2 & 8 & 0 & 2 & 8 & 0 & 1 & 9 & 0 & 0 & 10 & 0 & 1 & 9 \\
\midrule
\%Err & \multicolumn{3}{c|}{33.3} & \multicolumn{3}{c|}{26.7} & \multicolumn{3}{c|}{33.3} & \multicolumn{3}{c|}{33.3} & \multicolumn{3}{c|}{26.7} & \multicolumn{3}{c|}{20.0} & \multicolumn{3}{c|}{20.0} \\
\%Maj. Err & \multicolumn{3}{c|}{0.0} & \multicolumn{3}{c|}{0.0} & \multicolumn{3}{c|}{0.0} & \multicolumn{3}{c|}{0.0} & \multicolumn{3}{c|}{0.0} & \multicolumn{3}{c|}{0.0} & \multicolumn{3}{c|}{0.0} \\
\bottomrule
\end{tabular}
\caption[\textbf{ExpertQA: Inconsistency in system-pair ranking based on claim-level error rates for systems}]{\textbf{ExpertQA: Inconsistency in system-pair ranking based on claim-level error rates for systems.} We report a confusion matrix of pairwise system ranking decisions. We measure inconsistencies between the ranking based on the labeled error rate and the ranking based on the predicted error rate. For a system pair (s1, s2),  `=' indicates no significant difference between s1 and s2, `<' indicates s1 has a lower error rate than s2, and `>' indicates s1 has a higher error rate than s2. When a metric predicts a significant but opposite ranking between a pair, we count it as a
Major Error. Significance is computed with the two-proportion z-test and p\_value < 0.05.}
\label{tab:expertqa-instance-ranking-error}
\end{table*}

\begin{table*}
\centering
\tiny
\begin{tabular}{llllllll}
\toprule
 & label & GPT-4-turbo & GPT-3.5-turbo & Bespoke-7B & Bespoke-7B (cs=500) & MiniCheck-FT5 & MiniCheck-Roberta \\
System Name &  &  &  &  &  &  &  \\
\midrule
bing\_chat & 29.0 (0.0) & 54.4 (25.4) & 60.4 (31.4) & 63.3 (34.3) & 71.6 (42.6) & 73.4 (44.4) & 81.7 (52.7) \\
gpt4 & 39.2 (0.0) & 74.5 (35.3) & 70.6 (31.4) & 86.3 (47.1) & 82.4 (43.1) & 88.2 (49.0) & 94.1 (54.9) \\
post\_hoc\_gs\_gpt4 & 52.0 (0.0) & 91.3 (39.3) & 92.9 (40.8) & 98.0 (45.9) & 98.0 (45.9) & 98.0 (45.9) & 98.5 (46.4) \\
post\_hoc\_sphere\_gpt4 & 60.5 (0.0) & 86.8 (26.3) & 87.9 (27.4) & 94.2 (33.7) & 94.2 (33.7) & 94.7 (34.2) & 98.9 (38.4) \\
rr\_gs\_gpt4 & 26.6 (0.0) & 27.1 (0.5) & 33.0 (6.4) & 44.3 (17.7) & 44.8 (18.2) & 56.2 (29.6) & 63.1 (36.5) \\
rr\_sphere\_gpt4 & 42.1 (0.0) & 26.4 (-15.7) & 44.3 (2.1) & 45.0 (2.9) & 45.0 (2.9) & 61.4 (19.3) & 79.3 (37.1) \\
\midrule
Headroom & 26.6 (0.0) & 26.4 (-0.2) & 33.0 (6.4) & 44.3 (17.7) & 44.8 (18.2) & 56.2 (29.6) & 63.1 (36.5) \\
\bottomrule
\end{tabular}
\caption{\textbf{ExpertQA: Predicted summary-level error rates for systems.} Quantification bias in paratheses.}
\label{tab:expertqa-system-summary-error}
\end{table*}

\begin{table*}
\centering
\tiny
\begin{tabular}{llllllll}
\toprule
 & label & GPT-4-turbo & GPT-3.5-turbo & Bespoke-7B & Bespoke-7B (cs=500) & MiniCheck-FT5 & MiniCheck-Roberta \\
System Name &  &  &  &  &  &  &  \\
\midrule
alpaca & 76.0 (0.0) & 68.5 (-7.5) & 71.5 (-4.5) & 70.8 (-5.2) & 70.8 (-5.2) & 75.3 (-0.7) & 83.5 (7.5) \\
alpaca\_wdoc & 41.8 (0.0) & 34.7 (-7.0) & 44.6 (2.8) & 36.8 (-4.9) & 37.5 (-4.2) & 38.2 (-3.5) & 48.8 (7.0) \\
gpt3 & 78.9 (0.0) & 62.7 (-16.2) & 60.5 (-18.4) & 69.1 (-9.8) & 68.9 (-10.0) & 66.4 (-12.5) & 81.1 (2.3) \\
gpt3\_wdoc & 18.1 (0.0) & 15.8 (-2.3) & 22.3 (4.3) & 17.2 (-0.9) & 18.3 (0.3) & 22.6 (4.6) & 28.7 (10.6) \\
gpt3\_whudoc & 28.8 (0.0) & 20.5 (-8.3) & 25.1 (-3.7) & 25.1 (-3.7) & 25.6 (-3.1) & 30.5 (1.7) & 38.2 (9.4) \\
webgpt & 7.4 (0.0) & 6.5 (-0.9) & 13.9 (6.5) & 6.5 (-0.9) & 6.5 (-0.9) & 7.4 (0.0) & 9.6 (2.2) \\
\midrule
Headroom & 7.4 (0.0) & 6.5 (-0.9) & 13.9 (6.5) & 6.5 (-0.9) & 6.5 (-0.9) & 7.4 (0.0) & 9.6 (2.2) \\
\bottomrule
\end{tabular}
\caption{\textbf{LFQA: Predicted claim-level error rates for systems.} Quantification bias in paratheses.}
\label{tab:lfqa-system-instance-error}
\end{table*}

\begin{table*}
\centering
\tiny
\begin{tabular}{l|ccc|ccc|ccc|ccc|ccc|ccc|ccc|}
\toprule
GT Order & \multicolumn{3}{c|}{GPT-4-turbo} & \multicolumn{3}{c|}{GPT-3.5-turbo} & \multicolumn{3}{c|}{Bespoke-7B} & \multicolumn{3}{c|}{Bespoke-7B (cs=500)} & \multicolumn{3}{c|}{MiniCheck-FT5} & \multicolumn{3}{c|}{MiniCheck-Roberta} & \multicolumn{3}{c|}{AlignScore} \\
 & > & = & < & > & = & < & > & = & < & > & = & < & > & = & < & > & = & < & > & = & < \\
\midrule
= & 0 & 1 & 0 & 1 & 0 & 0 & 0 & 1 & 0 & 0 & 1 & 0 & 1 & 0 & 0 & 0 & 1 & 0 & 0 & 1 & 0 \\
< & 0 & 1 & 13 & 0 & 1 & 13 & 0 & 0 & 14 & 0 & 0 & 14 & 0 & 0 & 14 & 0 & 0 & 14 & 0 & 1 & 13 \\
\midrule
\%Err & \multicolumn{3}{c|}{6.7} & \multicolumn{3}{c|}{13.3} & \multicolumn{3}{c|}{0.0} & \multicolumn{3}{c|}{0.0} & \multicolumn{3}{c|}{6.7} & \multicolumn{3}{c|}{0.0} & \multicolumn{3}{c|}{6.7} \\
\%Maj. Err & \multicolumn{3}{c|}{0.0} & \multicolumn{3}{c|}{0.0} & \multicolumn{3}{c|}{0.0} & \multicolumn{3}{c|}{0.0} & \multicolumn{3}{c|}{0.0} & \multicolumn{3}{c|}{0.0} & \multicolumn{3}{c|}{0.0} \\
\bottomrule
\end{tabular}
\caption[\textbf{LFQA: Inconsistency in system-pair ranking based on claim-level error rates for systems}]{\textbf{LFQA: Inconsistency in system-pair ranking based on claim-level error rates for systems.} We report a confusion matrix of pairwise system ranking decisions. We measure inconsistencies between the ranking based on the labeled error rate and the ranking based on the predicted error rate. For a system pair (s1, s2),  `=' indicates no significant difference between s1 and s2, `<' indicates s1 has a lower error rate than s2, and `>' indicates s1 has a higher error rate than s2. When a metric predicts a significant but opposite ranking between a pair, we count it as a
Major Error. Significance is computed with the two-proportion z-test and p\_value < 0.05.}
\label{tab:lfqa-instance-ranking-error}
\end{table*}

\begin{table*}
\centering
\tiny
\begin{tabular}{llllllll}
\toprule
 & label & GPT-4-turbo & GPT-3.5-turbo & Bespoke-7B & Bespoke-7B (cs=500) & MiniCheck-FT5 & MiniCheck-Roberta \\
System Name &  &  &  &  &  &  &  \\
\midrule
alpaca & 100.0 (0.0) & 96.0 (-4.0) & 100.0 (0.0) & 100.0 (0.0) & 100.0 (0.0) & 100.0 (0.0) & 100.0 (0.0) \\
alpaca\_wdoc & 72.0 (0.0) & 68.0 (-4.0) & 90.0 (18.0) & 80.0 (8.0) & 84.0 (12.0) & 72.0 (0.0) & 76.0 (4.0) \\
gpt3 & 100.0 (0.0) & 100.0 (0.0) & 100.0 (0.0) & 100.0 (0.0) & 100.0 (0.0) & 100.0 (0.0) & 100.0 (0.0) \\
gpt3\_wdoc & 56.0 (0.0) & 56.0 (0.0) & 68.0 (12.0) & 56.0 (0.0) & 62.0 (6.0) & 68.0 (12.0) & 78.0 (22.0) \\
gpt3\_whudoc & 68.0 (0.0) & 60.0 (-8.0) & 72.0 (4.0) & 64.0 (-4.0) & 64.0 (-4.0) & 78.0 (10.0) & 86.0 (18.0) \\
webgpt & 36.0 (0.0) & 32.0 (-4.0) & 52.0 (16.0) & 26.0 (-10.0) & 26.0 (-10.0) & 32.0 (-4.0) & 38.0 (2.0) \\
\midrule
Headroom & 36.0 (0.0) & 32.0 (-4.0) & 52.0 (16.0) & 26.0 (-10.0) & 26.0 (-10.0) & 32.0 (-4.0) & 38.0 (2.0) \\
\bottomrule
\end{tabular}
\caption{\textbf{LFQA: Predicted summary-level error rates for systems.} Quantification bias in paratheses.}
\label{tab:lfqa-system-summary-error}
\end{table*}

\begin{table*}
\centering
\tiny
\begin{tabular}{lllllllll}
\toprule
 &  & GT Error Rate & GPT-4-turbo & GPT-3.5-turbo & Bespoke-7B & Bespoke-7B (cs=500) & MiniCheck-FT5 & MiniCheck-Roberta \\
Query Set & System Name &  &  &  &  &  &  &  \\
\midrule
\multirow[t]{7}{*}{CNN/DM} & gpt-3.5-turbo-0613 & 0.8 (0.0) & 1.5 (0.7) & 4.9 (4.1) & 2.1 (1.2) & 5.6 (4.8) & 7.0 (6.2) & 7.7 (6.9) \\
 & gpt-4-0613 & 1.9 (0.0) & 1.6 (-0.2) & 5.1 (3.3) & 4.9 (3.0) & 8.4 (6.5) & 7.4 (5.6) & 7.4 (5.6) \\
 & llama-2-70b-chat & 4.8 (0.0) & 7.1 (2.3) & 9.5 (4.6) & 10.9 (6.1) & 18.9 (14.1) & 16.6 (11.8) & 26.3 (21.4) \\
 & llama-2-13b-chat & 9.6 (0.0) & 12.2 (2.6) & 10.8 (1.2) & 15.7 (6.1) & 25.9 (16.3) & 25.4 (15.7) & 30.6 (21.0) \\
 & llama-2-7b-chat & 13.5 (0.0) & 17.1 (3.6) & 13.5 (0.0) & 17.9 (4.4) & 27.5 (14.0) & 25.6 (12.2) & 33.2 (19.7) \\
 & mistral-7B-instruct & 13.5 (0.0) & 17.4 (3.9) & 17.8 (4.3) & 16.2 (2.7) & 21.1 (7.6) & 19.5 (5.9) & 25.4 (11.9) \\
 \cline{2-9}
 & Headroom & 0.8 (0.0) & 1.5 (0.7) & 4.9 (4.1) & 2.1 (1.2) & 5.6 (4.8) & 7.0 (6.2) & 7.4 (6.6) \\
\cline{1-9}
\multirow[t]{7}{*}{Recent News} & gpt-3.5-turbo-0613 & 0.8 (0.0) & 1.7 (0.8) & 8.0 (7.2) & 2.5 (1.7) & 4.6 (3.8) & 3.8 (3.0) & 7.2 (6.3) \\
 & gpt-4-0613 & 1.9 (0.0) & 3.3 (1.4) & 10.0 (8.1) & 2.9 (1.0) & 4.3 (2.4) & 8.6 (6.7) & 12.4 (10.5) \\
 & llama-2-70b-chat & 5.4 (0.0) & 5.9 (0.5) & 13.4 (7.9) & 7.9 (2.5) & 10.9 (5.4) & 13.9 (8.4) & 21.8 (16.3) \\
 & llama-2-13b-chat & 10.3 (0.0) & 12.0 (1.7) & 17.9 (7.7) & 16.2 (6.0) & 18.8 (8.5) & 17.9 (7.7) & 32.5 (22.2) \\
 & llama-2-7b-chat & 11.1 (0.0) & 20.1 (9.0) & 21.5 (10.4) & 16.7 (5.6) & 18.8 (7.6) & 20.1 (9.0) & 38.2 (27.1) \\
 & mistral-7B-instruct & 16.2 (0.0) & 18.4 (2.1) & 19.7 (3.4) & 15.0 (-1.3) & 18.4 (2.1) & 15.8 (-0.4) & 23.5 (7.3) \\
 \cline{2-9}
 & Headroom & 0.8 (0.0) & 1.7 (0.8) & 8.0 (7.2) & 2.5 (1.7) & 4.3 (3.5) & 3.8 (3.0) & 7.2 (6.3) \\
\cline{1-9}
\multirow[t]{7}{*}{MARCO} & gpt-3.5-turbo-0613 & 1.9 (0.0) & 6.5 (4.7) & 14.2 (12.3) & 8.0 (6.2) & 8.2 (6.3) & 8.2 (6.3) & 11.4 (9.5) \\
 & gpt-4-0613 & 0.6 (0.0) & 3.2 (2.6) & 13.9 (13.4) & 4.3 (3.8) & 4.6 (4.0) & 7.5 (7.0) & 6.7 (6.1) \\
 & llama-2-70b-chat & 3.6 (0.0) & 21.0 (17.4) & 28.1 (24.5) & 26.2 (22.6) & 26.0 (22.4) & 27.8 (24.2) & 32.3 (28.7) \\
 & llama-2-13b-chat & 7.0 (0.0) & 22.8 (15.8) & 30.5 (23.5) & 24.5 (17.5) & 24.8 (17.8) & 25.1 (18.1) & 30.5 (23.5) \\
 & llama-2-7b-chat & 7.0 (0.0) & 26.8 (19.8) & 33.1 (26.1) & 27.6 (20.6) & 27.8 (20.8) & 27.4 (20.4) & 33.7 (26.7) \\
 & mistral-7B-instruct & 8.4 (0.0) & 23.4 (15.0) & 31.9 (23.4) & 23.9 (15.5) & 23.9 (15.5) & 24.5 (16.1) & 26.2 (17.8) \\
 \cline{2-9}
 & Headroom & 0.6 (0.0) & 3.2 (2.6) & 13.9 (13.4) & 4.3 (3.8) & 4.6 (4.0) & 7.5 (7.0) & 6.7 (6.1) \\
\cline{1-9}
\multirow[t]{7}{*}{Yelp} & gpt-3.5-turbo-0613 & 2.7 (0.0) & 3.1 (0.4) & 16.2 (13.5) & 7.0 (4.3) & 12.5 (9.8) & 37.1 (34.4) & 24.8 (22.1) \\
 & gpt-4-0613 & 3.5 (0.0) & 1.5 (-2.0) & 23.6 (20.1) & 9.9 (6.4) & 17.9 (14.4) & 57.7 (54.2) & 31.9 (28.4) \\
 & llama-2-70b-chat & 19.5 (0.0) & 28.5 (9.0) & 46.2 (26.7) & 50.8 (31.2) & 58.9 (39.4) & 67.7 (48.2) & 58.1 (38.6) \\
 & llama-2-13b-chat & 26.7 (0.0) & 31.9 (5.2) & 45.3 (18.6) & 46.7 (20.0) & 57.0 (30.3) & 68.9 (42.2) & 60.6 (33.9) \\
 & llama-2-7b-chat & 24.5 (0.0) & 29.0 (4.5) & 47.6 (23.1) & 46.7 (22.2) & 56.7 (32.3) & 66.5 (42.0) & 55.3 (30.8) \\
 & mistral-7B-instruct & 21.7 (0.0) & 24.5 (2.8) & 35.0 (13.3) & 29.7 (8.0) & 37.0 (15.4) & 56.8 (35.1) & 38.3 (16.6) \\
 \cline{2-9}
 & Headroom & 2.7 (0.0) & 1.5 (-1.1) & 16.2 (13.5) & 7.0 (4.3) & 12.5 (9.8) & 37.1 (34.4) & 24.8 (22.1) \\
\bottomrule
\end{tabular}
\caption{\textbf{RAGTruth: Predicted claim-level error rates for systems.} Quantification bias in paratheses.}
\label{tab:ragtruth-system-instance-error}
\end{table*}

\begin{table*}
\centering
\tiny
\begin{tabular}{l|ccc|ccc|ccc|ccc|ccc|ccc|ccc|}
\toprule
GT Order & \multicolumn{3}{c|}{GPT-4-turbo} & \multicolumn{3}{c|}{GPT-3.5-turbo} & \multicolumn{3}{c|}{Bespoke-7B} & \multicolumn{3}{c|}{Bespoke-7B (cs=500)} & \multicolumn{3}{c|}{MiniCheck-FT5} & \multicolumn{3}{c|}{MiniCheck-Roberta} & \multicolumn{3}{c|}{AlignScore} \\
 & > & = & < & > & = & < & > & = & < & > & = & < & > & = & < & > & = & < & > & = & < \\
\midrule
\multicolumn{22}{c}{\textbf{RAGTruth-CNN/DM}} \\
\midrule
= & 0 & 2 & 1 & 0 & 2 & 1 & 0 & 3 & 0 & 1 & 2 & 0 & 2 & 1 & 0 & 1 & 2 & 0 & 0 & 3 & 0 \\
< & 0 & 0 & 3 & 0 & 2 & 1 & 0 & 0 & 3 & 0 & 1 & 2 & 0 & 1 & 2 & 0 & 2 & 1 & 0 & 1 & 2 \\
\midrule
\%Err & \multicolumn{3}{c|}{16.7} & \multicolumn{3}{c|}{50.0} & \multicolumn{3}{c|}{0.0} & \multicolumn{3}{c|}{33.3} & \multicolumn{3}{c|}{50.0} & \multicolumn{3}{c|}{50.0} & \multicolumn{3}{c|}{16.7}\\
\%Maj. Err & \multicolumn{3}{c|}{0.0} & \multicolumn{3}{c|}{0.0} & \multicolumn{3}{c|}{0.0} & \multicolumn{3}{c|}{0.0} & \multicolumn{3}{c|}{0.0} & \multicolumn{3}{c|}{0.0} & \multicolumn{3}{c|}{0.0} \\

\midrule
\multicolumn{22}{c}{\textbf{RAGTruth-News}} \\
\midrule
= & 0 & 4 & 1 & 0 & 4 & 1 & 0 & 3 & 2 & 0 & 3 & 2 & 0 & 5 & 0 & 1 & 2 & 2 & 0 & 4 & 1 \\
< & 0 & 0 & 1 & 0 & 1 & 0 & 0 & 0 & 1 & 0 & 0 & 1 & 0 & 1 & 0 & 0 & 1 & 0 & 0 & 0 & 1 \\
\midrule
\%Err & \multicolumn{3}{c|}{16.7} & \multicolumn{3}{c|}{33.3} & \multicolumn{3}{c|}{33.3} & \multicolumn{3}{c|}{33.3} & \multicolumn{3}{c|}{16.7} & \multicolumn{3}{c|}{66.7} & \multicolumn{3}{c|}{16.7} \\
\%Maj. Err & \multicolumn{3}{c|}{0.0} & \multicolumn{3}{c|}{0.0} & \multicolumn{3}{c|}{0.0} & \multicolumn{3}{c|}{0.0} & \multicolumn{3}{c|}{0.0} & \multicolumn{3}{c|}{0.0} & \multicolumn{3}{c|}{0.0} \\

\midrule
\multicolumn{22}{c}{\textbf{RAGTruth-MARCO}} \\
\midrule
= & 0 & 2 & 1 & 0 & 3 & 0 & 0 & 3 & 0 & 0 & 3 & 0 & 0 & 3 & 0 & 1 & 2 & 0 & 1 & 2 & 0 \\
< & 0 & 2 & 1 & 0 & 2 & 1 & 0 & 3 & 0 & 0 & 3 & 0 & 0 & 3 & 0 & 1 & 2 & 0 & 0 & 3 & 0 \\
\midrule
\%Err & \multicolumn{3}{c|}{50.0} & \multicolumn{3}{c|}{33.3} & \multicolumn{3}{c|}{50.0} & \multicolumn{3}{c|}{50.0} & \multicolumn{3}{c|}{50.0} & \multicolumn{3}{c|}{66.7} & \multicolumn{3}{c|}{66.7} \\
\%Maj. Err & \multicolumn{3}{c|}{0.0} & \multicolumn{3}{c|}{0.0} & \multicolumn{3}{c|}{0.0} & \multicolumn{3}{c|}{0.0} & \multicolumn{3}{c|}{0.0} & \multicolumn{3}{c|}{16.7} & \multicolumn{3}{c|}{0.0} \\

\midrule
\multicolumn{22}{c}{\textbf{RAGTruth-Yelp}} \\
\midrule
= & 2 & 1 & 1 & 1 & 1 & 2 & 1 & 1 & 2 & 1 & 1 & 2 & 1 & 1 & 2 & 1 & 0 & 3 & 1 & 2 & 1 \\
< & 0 & 2 & 9 & 0 & 2 & 9 & 0 & 2 & 9 & 0 & 2 & 9 & 0 & 3 & 8 & 0 & 2 & 9 & 0 & 2 & 9 \\
\midrule
\%Err & \multicolumn{3}{c|}{33.3} & \multicolumn{3}{c|}{33.3} & \multicolumn{3}{c|}{33.3} & \multicolumn{3}{c|}{33.3} & \multicolumn{3}{c|}{40.0} & \multicolumn{3}{c|}{40.0} & \multicolumn{3}{c|}{26.7} \\
\%Maj. Err & \multicolumn{3}{c|}{0.0} & \multicolumn{3}{c|}{0.0} & \multicolumn{3}{c|}{0.0} & \multicolumn{3}{c|}{0.0} & \multicolumn{3}{c|}{0.0} & \multicolumn{3}{c|}{0.0} & \multicolumn{3}{c|}{0.0} \\

\bottomrule
\end{tabular}
\caption[\textbf{RAGTruth: Inconsistency in system-pair ranking based on claim-level error rates for systems}]{\textbf{RAGTruth: Inconsistency in system-pair ranking based on claim-level error rates for systems.} We report a confusion matrix of pairwise system ranking decisions. We measure inconsistencies between the ranking based on the labeled error rate and the ranking based on the predicted error rate. For a system pair (s1, s2),  `=' indicates no significant difference between s1 and s2, `<' indicates s1 has a lower error rate than s2, and `>' indicates s1 has a higher error rate than s2. When a metric predicts a significant but opposite ranking between a pair, we count it as a
Major Error. Significance is computed with the two-proportion z-test and p\_value < 0.05.}
\label{tab:ragtruth-instance-ranking-error}
\end{table*}

\begin{table*}
\centering
\tiny
\begin{tabular}{lllllllll}
\toprule
 &  & GT Error Rate & GPT-4-turbo & GPT-3.5-turbo & Bespoke-7B & Bespoke-7B (cs=500) & MiniCheck-FT5 & MiniCheck-Roberta \\
Query Set & System Name &  &  &  &  &  &  &  \\
\midrule
\multirow[t]{7}{*}{CNN/DM} & gpt-3.5-turbo-0613 & 1.5 (0.0) & 2.7 (1.2) & 8.8 (7.2) & 3.5 (2.0) & 9.3 (7.8) & 12.3 (10.8) & 13.5 (12.0) \\
 & gpt-4-0613 & 2.3 (0.0) & 2.3 (0.0) & 7.0 (4.7) & 6.7 (4.3) & 11.7 (9.4) & 10.0 (7.7) & 10.4 (8.0) \\
 & llama-2-13b-chat & 12.0 (0.0) & 14.7 (2.6) & 13.5 (1.5) & 19.2 (7.1) & 32.0 (19.9) & 31.6 (19.5) & 36.8 (24.8) \\
 & llama-2-70b-chat & 7.3 (0.0) & 10.7 (3.5) & 14.2 (6.9) & 16.1 (8.8) & 26.5 (19.2) & 23.0 (15.8) & 36.9 (29.7) \\
 & llama-2-7b-chat & 18.4 (0.0) & 23.5 (5.1) & 18.4 (0.0) & 24.2 (5.8) & 36.8 (18.4) & 33.9 (15.5) & 43.0 (24.5) \\
 & mistral-7B-instruct & 18.8 (0.0) & 24.7 (5.9) & 24.1 (5.3) & 22.5 (3.7) & 29.1 (10.3) & 27.2 (8.4) & 32.5 (13.7) \\
 \cline{2-9}
 & Headroom & 1.5 (0.0) & 2.3 (0.8) & 7.0 (5.5) & 3.5 (2.0) & 9.3 (7.8) & 10.0 (8.5) & 10.4 (8.9) \\
\cline{1-9}
\multirow[t]{6}{*}{Recent News} & gpt-3.5-turbo-0613 & 1.2 (0.0) & 2.5 (1.2) & 11.8 (10.6) & 3.7 (2.5) & 6.8 (5.6) & 5.6 (4.3) & 10.6 (9.3) \\
 & gpt-4-0613 & 2.6 (0.0) & 4.6 (2.0) & 13.7 (11.1) & 3.9 (1.3) & 5.9 (3.3) & 11.1 (8.5) & 16.3 (13.7) \\
 & llama-2-13b-chat & 11.8 (0.0) & 12.7 (1.0) & 20.6 (8.8) & 18.6 (6.9) & 21.6 (9.8) & 20.6 (8.8) & 35.3 (23.5) \\
 & llama-2-70b-chat & 7.5 (0.0) & 8.2 (0.7) & 17.8 (10.3) & 11.0 (3.4) & 14.4 (6.8) & 18.5 (11.0) & 27.4 (19.9) \\
 & llama-2-7b-chat & 12.8 (0.0) & 23.9 (11.1) & 25.6 (12.8) & 20.5 (7.7) & 23.1 (10.3) & 23.9 (11.1) & 44.4 (31.6) \\
 & mistral-7B-instruct & 23.8 (0.0) & 25.0 (1.2) & 26.2 (2.5) & 20.6 (-3.1) & 25.0 (1.2) & 21.9 (-1.9) & 31.9 (8.1) \\
 \cline{2-9}
 & Headroom & 1.2 (0.0) & 2.5 (1.2) & 11.8 (10.6) & 3.7 (2.5) & 5.9 (4.6) & 5.6 (4.3) & 10.6 (9.3) \\
\cline{1-9}
\multirow[t]{6}{*}{MARCO} & gpt-3.5-turbo-0613 & 2.8 (0.0) & 8.6 (5.8) & 17.7 (14.9) & 10.5 (7.7) & 10.8 (8.0) & 10.2 (7.5) & 14.4 (11.6) \\
 & gpt-4-0613 & 0.8 (0.0) & 4.8 (4.0) & 18.8 (17.9) & 6.5 (5.6) & 6.9 (6.0) & 10.6 (9.8) & 9.6 (8.8) \\
 & llama-2-13b-chat & 11.2 (0.0) & 30.7 (19.5) & 40.2 (29.0) & 33.0 (21.8) & 33.4 (22.2) & 35.1 (23.9) & 40.6 (29.4) \\
 & llama-2-70b-chat & 6.1 (0.0) & 29.7 (23.6) & 36.8 (30.8) & 34.7 (28.7) & 34.5 (28.5) & 37.4 (31.4) & 45.0 (38.9) \\
 & llama-2-7b-chat & 11.3 (0.0) & 37.3 (26.0) & 43.8 (32.5) & 37.2 (25.9) & 37.5 (26.2) & 38.4 (27.1) & 45.8 (34.5) \\
 & mistral-7B-instruct & 11.1 (0.0) & 30.1 (19.0) & 40.3 (29.1) & 30.6 (19.4) & 30.6 (19.4) & 32.5 (21.3) & 34.6 (23.5) \\
 \cline{2-9}
 & Headroom & 0.8 (0.0) & 4.8 (4.0) & 17.7 (16.8) & 6.5 (5.6) & 6.9 (6.0) & 10.2 (9.4) & 9.6 (8.8) \\
\cline{1-9}
\multirow[t]{6}{*}{Yelp} & gpt-3.5-turbo-0613 & 5.7 (0.0) & 7.1 (1.4) & 32.1 (26.4) & 15.5 (9.8) & 26.5 (20.8) & 59.7 (54.0) & 46.9 (41.1) \\
 & gpt-4-0613 & 5.8 (0.0) & 2.6 (-3.2) & 35.1 (29.3) & 16.3 (10.5) & 27.4 (21.6) & 72.3 (66.5) & 46.6 (40.9) \\
 & llama-2-13b-chat & 37.9 (0.0) & 44.7 (6.8) & 59.0 (21.1) & 59.0 (21.1) & 68.7 (30.8) & 77.6 (39.6) & 72.9 (35.0) \\
 & llama-2-70b-chat & 29.8 (0.0) & 41.5 (11.7) & 59.6 (29.8) & 64.2 (34.4) & 73.5 (43.7) & 78.0 (48.2) & 69.7 (39.9) \\
 & llama-2-7b-chat & 34.1 (0.0) & 39.9 (5.7) & 58.8 (24.6) & 58.2 (24.1) & 69.2 (35.0) & 76.9 (42.8) & 67.8 (33.6) \\
 & mistral-7B-instruct & 36.0 (0.0) & 39.9 (3.9) & 51.9 (15.9) & 46.7 (10.7) & 54.6 (18.6) & 72.9 (36.9) & 55.2 (19.2) \\
 \cline{2-9}
 & Headroom & 5.7 (0.0) & 2.6 (-3.1) & 32.1 (26.4) & 15.5 (9.8) & 26.5 (20.8) & 59.7 (54.0) & 46.6 (40.9) \\
\cline{1-9}
\bottomrule
\end{tabular}
\caption{\textbf{RAGTruth: Predicted summary-level error rates for systems.} Quantification bias in paratheses.}
\label{tab:ragtruth-system-summary-error}
\end{table*}

\subsection{Visualization of System-level Quantification Bias on RAGTruth}
\label{app:ragtruth-quantification-error}

\begin{figure}[tb]
    \centering
    \includegraphics[width=0.48\textwidth]{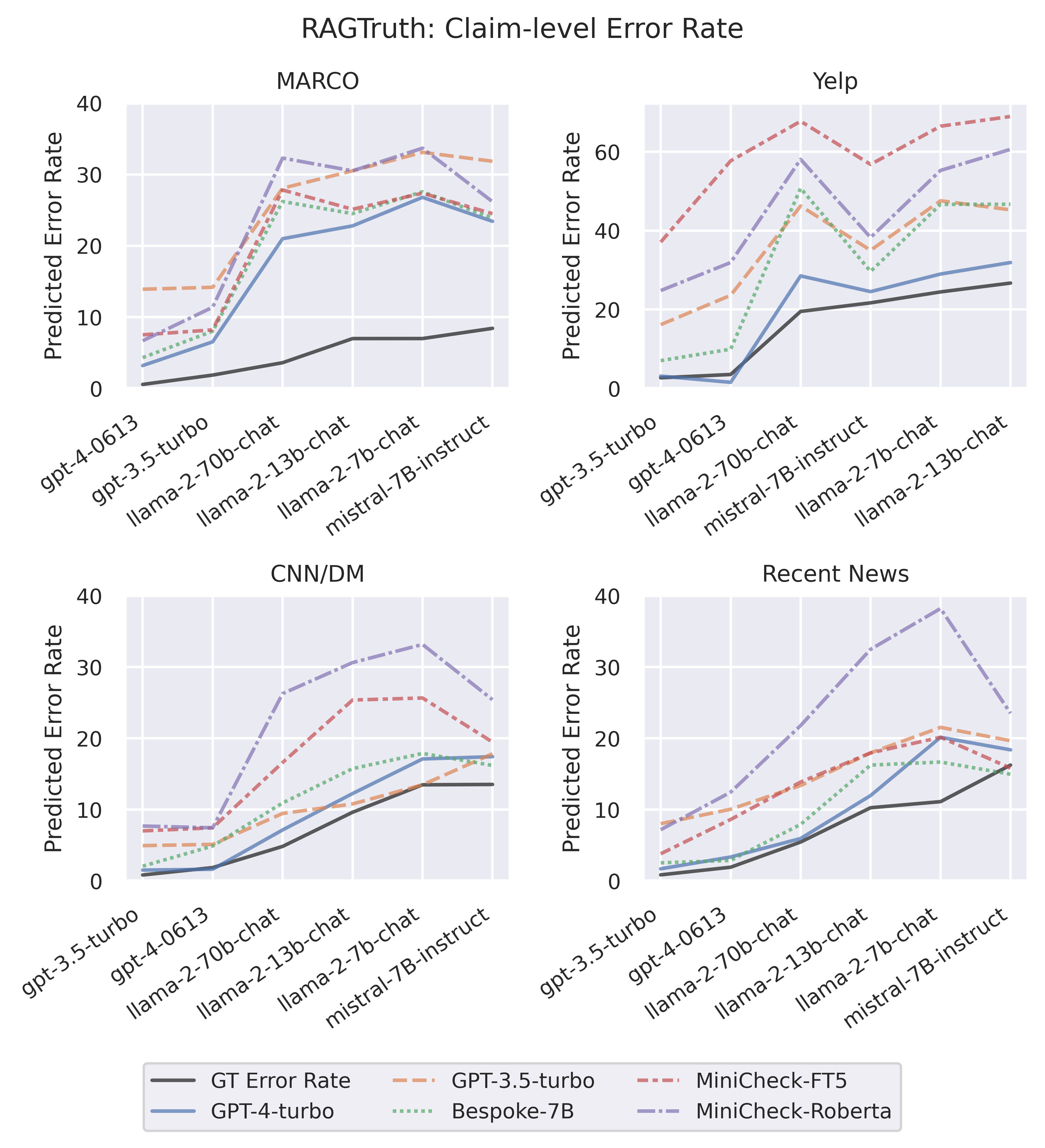}
    \caption{\textbf{Predicted system-level error rate on RAGTruth (claim-level).} Inconsistent predictions between different metrics lead to discrepancies in the quantification of the system error rate.
    }
    \label{fig:ragtruth-sys-err-rate}
\end{figure}

\begin{figure}[tb]
    \centering
    \includegraphics[width=0.48\textwidth]{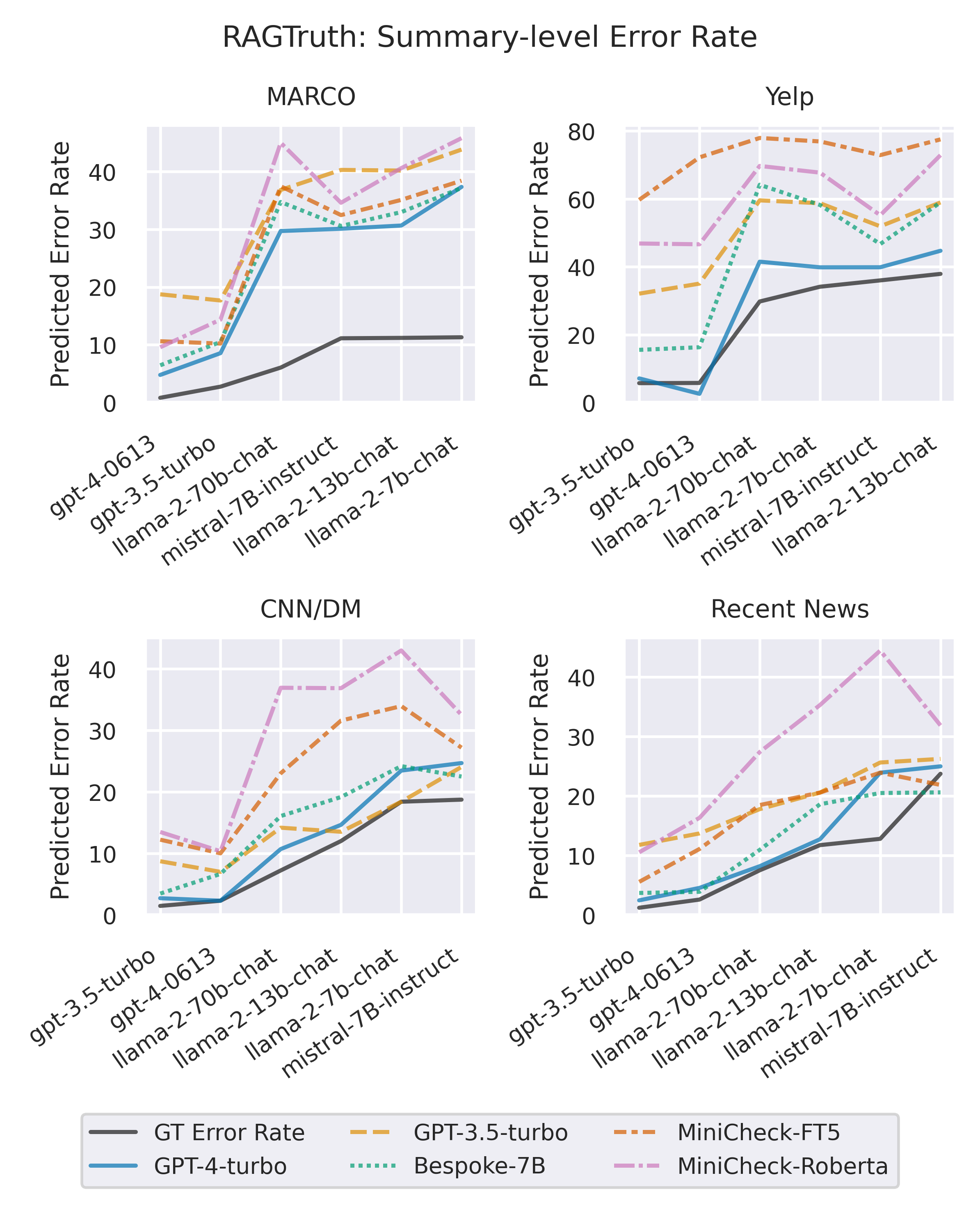}
    \caption{\textbf{Predicted system-level error rate on RAGTruth (summary-level).} Claim-level misclassification and metric inconsistency lead to even larger summary-level quantification bias.
    }
    \label{fig:ragtruth-summ-sys-err-rate}
\end{figure}

In Figure~\ref{fig:ragtruth-sys-err-rate}, we highlight the bias of the metrics in predicting the claim-level error rate on the RAGTruth dataset. We see that the bias of the top AutoAIS metrics is consistently poor on the MSMARCO subset, especially on the systems with a higher ground-truth hallucination rate (e.g. the bias is 15-20\% for \bespoke). On the Yelp subset, we see that all metrics besides \gptf show poor ground truth error estimation; the bias of \gptf is 3.6\% (in magnitude) on average as opposed to 13.8\% (in magnitude) for \bespoke. This is especially glaring since balanced accuracy does not indicate a large difference between \gptf and \bespoke (84.7\% BAcc vs 81.6\% BAcc). On the summarization subsets of RAGTruth (CNN-DM and Recent News), we see that the metrics predict large differences between systems when the ground-truth annotation does not and vice versa. For example, while ground truth annotations predict that Llama-2-13B-chat makes much fewer grounding errors than Mistral-7B-Instruct (9.6\% vs 13.5\%), \bespoke predicts Mistral-7B-Instruct to be on par with Llama-2-13B-chat. Thus, results indicate several inconsistencies between predicted and ground-truth system error rates. We report trends for response-level bias of the metrics in Figure~\ref{fig:ragtruth-summ-sys-err-rate}.

\subsection{Effect of Chunking on Evaluator}
\label{app:chunking}

In Table~\ref{tab:app-chunking-effect}, we report the performance of the \bespoke evaluator without and with chunking (chunk size of 500 words). We report the performance on the subset of examples where chunking is applicable, i.e., examples where the document was longer than 500 words. 

\begin{table}
\centering
\scriptsize
\begin{tabular}{p{52pt}p{40pt}cccc}
\toprule
Dataset & Evaluator & BAcc & PPR & TPR & TNR \\
\midrule
\multirow{2}{*}{AggreFact-CNN} & Bespoke-7B & 58.4 & 89.7 & 92.3 & 24.4 \\
 & + chunk(500) & 60.4 & 79.8 & 83.0 & 37.8 \\
\cmidrule(lr){1-6}
\multirow{2}{*}{AggreFact-XSum} & Bespoke-7B & 69.7 & 58.3 & 74.4 & 65.1 \\
 & + chunk(500) & 68.8 & 52.5 & 67.8 & 69.9 \\
\cmidrule(lr){1-6}
\multirow{2}{*}{TofuEval-MediaS} & Bespoke-7B & 72.1 & 82.9 & 91.6 & 52.5 \\
 & + chunk(500) & 72.0 & 75.2 & 83.8 & 60.2 \\
\cmidrule(lr){1-6}
\multirow{2}{*}{TofuEval-MeetB} & Bespoke-7B & 77.1 & 80.8 & 90.6 & 63.7 \\
 & + chunk(500) & 75.8 & 72.7 & 82.0 & 69.6 \\
\cmidrule(lr){1-6}
\multirow{2}{*}{RAGTruth-CNN} & Bespoke-7B & 77.4 & 90.0 & 93.4 & 61.4 \\
 & + chunk(500) & 77.8 & 82.6 & 86.0 & 69.7 \\
\cmidrule(lr){1-6}
\multirow{2}{*}{RAGTruth-News} & Bespoke-7B & 78.7 & 89.3 & 94.0 & 63.5 \\
 & + chunk(500) & 78.4 & 84.8 & 89.5 & 67.3 \\
\cmidrule(lr){1-6}
\multirow{2}{*}{ClaimVerify} & Bespoke-7B & 74.6 & 78.4 & 90.3 & 58.8 \\
 & + chunk(500) & 74.6 & 78.0 & 89.9 & 59.3 \\
\cmidrule(lr){1-6}
\multirow{2}{*}{Wice} & Bespoke-7B & 85.5 & 36.7 & 84.4 & 86.5 \\
 & + chunk(500) & 76.9 & 27.1 & 63.3 & 90.6 \\
\cmidrule(lr){1-6}
\multirow{2}{*}{ExpertQA} & Bespoke-7B & 61.9 & 61.2 & 65.2 & 58.7 \\
 & + chunk(500) & 60.5 & 54.9 & 58.4 & 62.7 \\
\cmidrule(lr){1-6}
\multirow{2}{*}{Lfqa} & Bespoke-7B & 81.6 & 67.5 & 94.3 & 68.9 \\
 & + chunk(500) & 80.7 & 66.0 & 92.0 & 69.4 \\
\cmidrule(lr){1-6}
\multirow{2}{*}{RAGTruth-MARCO} & Bespoke-7B & 85.9 & 83.7 & 86.0 & 85.7 \\
 & + chunk(500) & 85.3 & 82.6 & 84.8 & 85.7 \\
\cmidrule(lr){1-6}
\multirow{2}{*}{RAGTruth-Yelp} & Bespoke-7B & 81.8 & 71.6 & 80.9 & 82.7 \\
 & + chunk(500) & 78.7 & 63.1 & 71.5 & 85.9 \\
\bottomrule
\end{tabular}
\caption[\textbf{Change in \bespoke evaluator predictions with document chunking}]{\textbf{Change in \bespoke evaluator predictions with document chunking:} We report the performance of the \bespoke evaluator without and with input document chunking (chunk size of 500 words). These results are calculated on the subset of examples where chunking is applicable. The evaluator with chunking has a lower rate of predicting label "attributable" (PPR = percent of examples predicted as positive/attributable). Correspondingly, the TPR is lower, while TNR is higher.}
\label{tab:app-chunking-effect}
\end{table}

\subsection{Details of Metric Adjustment for Reducing Bias}
\label{app:adjustment}

\begin{table*}
\centering
\scriptsize
\begin{tabular}{llc|ccccc}
\toprule
Source & Calibration Model & GT Error Rate & No Adjustment & Adjusted Counts & Thres. tuning for zero bias & Thres. tuning for $\uparrow$BAcc \\
\midrule
\multirow{6}{*}{CNN/DM} & gpt-3.5-turbo-0613 & 0.8 & 4.5 (6.1) & 65.6 (86.5) & \cellcolor{green3}2.3 (4.7) & 37.0 (42.9) \\
& gpt-4-0613 & 1.9 & 4.1 (6.1) & 18.2 (27.5) & \cellcolor{green3}1.5 (3.5) & \cellcolor{green3}3.1 (5.7) \\
& llama-2-70b-chat & 4.8 & 3.5 (6.1) & \cellcolor{green3}2.1 (3.7) & \cellcolor{green3}2.2 (4.7) & 11.9 (17.8) \\
& llama-2-13b-chat & 9.6 & 3.5 (6.1) & \cellcolor{green3}1.8 (3.2) & \cellcolor{green3}1.6 (3.5) & 11.3 (16.3) \\
& llama-2-7b-chat & 13.5 & 3.8 (6.1) & \cellcolor{green3}2.4 (4.8) & \cellcolor{green3}1.6 (3.6) & 21.0 (27.7) \\
& mistral-7B-instruct & 13.5 & 4.2 (6.1) & \cellcolor{green3}1.8 (3.2) & \cellcolor{green3}2.0 (3.8) & 4.8 (7.3) \\

\midrule
\multirow{6}{*}{Recent News} & gpt-3.5-turbo-0613 & 0.8 & 3.3 (6.0) & 11.2 (19.3) & \cellcolor{green3}1.7 (3.4) & 7.7 (15.4) \\
& gpt-4-0613 & 1.9 & 3.4 (6.0) & 32.3 (52.0) & \cellcolor{green3}2.7 (4.3) & 13.1 (24.8) \\
& llama-2-70b-chat & 5.4 & 3.1 (6.0) & 8.4 (16.4) & \cellcolor{green3}1.6 (3.4) & 7.7 (15.4) \\
& llama-2-13b-chat & 10.3 & 2.4 (5.6) & 3.6 (9.4) & \cellcolor{green3}1.8 (5.6) & \cellcolor{green3}1.9 (4.2) \\
& llama-2-7b-chat & 11.1 & 2.5 (6.0) & 3.2 (7.7) & \cellcolor{green3}1.7 (5.6) & 12.3 (24.8) \\
& mistral-7B-instruct & 16.2 & 3.3 (6.0) & 4.3 (8.3) & 4.3 (7.7) & 18.7 (28.2) \\

\midrule
\multirow{6}{*}{MARCO} & gpt-4-0613 & 0.6 & 16.5 (22.6) & \cellcolor{green3}5.6 (8.4) & \cellcolor{green3}6.8 (10.4) & 38.4 (44.4) \\
& gpt-3.5-turbo-0613 & 1.9 & 16.0 (22.6) & 22.9 (32.6) & \cellcolor{green3}3.7 (7.2) & 30.0 (36.1) \\
& llama-2-70b-chat & 3.6 & 12.7 (20.6) & \cellcolor{green3}3.7 (8.4) & \cellcolor{green3}5.0 (8.4) & 17.8 (27.7) \\
& llama-2-13b-chat & 7.0 & 13.7 (22.6) & \cellcolor{green3}3.3 (6.4) & \cellcolor{green3}1.4 (4.7) & \cellcolor{green3}6.8 (12.4) \\
& llama-2-7b-chat & 7.0 & 13.1 (22.6) & \cellcolor{green3}3.6 (8.4) & \cellcolor{green3}1.6 (4.7) & 14.3 (24.0) \\
& mistral-7B-instruct & 8.4 & 14.1 (22.6) & \cellcolor{green3}3.9 (8.2) & \cellcolor{green3}1.9 (4.7) & 14.1 (22.6) \\

\midrule
\multirow{6}{*}{Yelp} & gpt-3.5-turbo-0613 & 2.7 & 17.6 (31.2) & 52.8 (80.5) & \cellcolor{green3}6.7 (16.2) & 32.9 (46.7) \\
& gpt-4-0613 & 3.5 & 17.1 (31.2) & 62.1 (80.5) & \cellcolor{green3}8.3 (19.4) & 53.5 (66.9) \\
& llama-2-70b-chat & 19.5 & 12.2 (22.2) & \cellcolor{green3}11.4 (21.7) & \cellcolor{green3}6.6 (10.7) & \cellcolor{green3}4.5 (11.3) \\
& mistral-7B-instruct & 21.7 & 16.8 (31.2) & 17.7 (35.9) & \cellcolor{green3}6.6 (16.2) & \cellcolor{green3}13.7 (26.7) \\
& llama-2-7b-chat & 24.5 & 14.0 (31.2) & \cellcolor{green3}8.3 (21.7) & \cellcolor{green3}4.0 (9.3) & \cellcolor{green3}6.1 (19.4) \\
& llama-2-13b-chat & 26.7 & 14.4 (31.2) & \cellcolor{green3}8.9 (21.7) & \cellcolor{green3}4.0 (6.7) & \cellcolor{green3}5.2 (16.2) \\
\bottomrule
\end{tabular}
\caption[\textbf{Comparison of adjustment methods on RAGTruth}]{\textbf{Comparison of adjustment methods on RAGTruth:} We report the bias in estimating the ground-truth system error (hallucination) rates using three adjustment methods.
In each section, we report mean absolute bias by using one system for calibration and calculating the mean absolute bias over the remaining systems. Numbers in parentheses indicate the worst-case bias over the remaining systems. \colorbox{green3}{Green cells} indicate a decrease in bias relative to "No Adjustment". Tuning the evaluator threshold for zero bias consistently reduces the absolute bias in estimation over the held-out systems. Threshold tuning to maximize BAcc worsens the estimation of system-level error. We see that the adjusted counts approach leads to high mean absolute bias when the ground truth error rate of the system is low.}
\label{tab:app-ragtruth-metric-adjustment}
\end{table*}

We compare three ways to reduce the bias of \ais evaluators in estimating the error rates of systems. \textbf{Adjusted Counts} ~\citep{10.1145/1150402.1150423-adjust-count} uses the TPR and FPR of the evaluator to adjust the predicted system level error rate ($\hat{p}_0$).
\begin{equation*}
    \hat{p} = \text{clip}(\frac{\hat{p}_0 - FPR}{TPR - FPR}, \text{min}=0, \text{max}=1)
\end{equation*}
Under this setup, we are estimating the prevalence (quantification) of hallucinations ($\hat{p}$) by extrapolating from the hallucination rate on a sample \citep{10.1145/3117807-review-on-quantification}). For our experiments, we compute the TPR and FPR of the \ais evaluator on the labeled claim-document pairs generated by one system and use it to adjust the predicted error rate ($\hat{p}_0$) of generations by the other systems. This method is appealing because it does not require the evaluator to produce a scalar score, i.e., it works with the predicted 0/1 labels.

When the evaluator predicts a score instead of directly predicting a label, we can apply threshold tuning. Same as before, we use the labeled claim-document pairs for one system to tune the threshold and then predict labels for the remaining held-out systems using this tuned threshold. We experiment with two tuning objectives: minimizing the absolute bias towards zero on the labeled calibration data or maximizing the BAcc on the labeled calibration data.

Table~\ref{tab:app-ragtruth-metric-adjustment} provides the resulting mean absolute bias by using each of the 6 systems one by one for calibration and computing bias on the remaining 5 systems. We report the average over all the calibration systems as the cross-validated bias in Table~\ref{tab:ragtruth-metric-adjustment}. We find that tuning the threshold for zero bias leads to consistent improvements in the held-out systems. Moreover, tuning for higher balanced accuracy hurts the error estimation on the held-out systems. We find that the adjusted counts approach does not provide an improvement over no adjustment if the system used for calibration has a low ground truth error rate. We believe that this is due to a skewed estimation of TPR and FPR when the prevalence of the label 0 is low.

\subsection{Claim-level Consistency of Metrics}
\label{app:metric-consistency}

\begin{figure}[htb]
    \centering
    \includegraphics[width=0.48\textwidth]{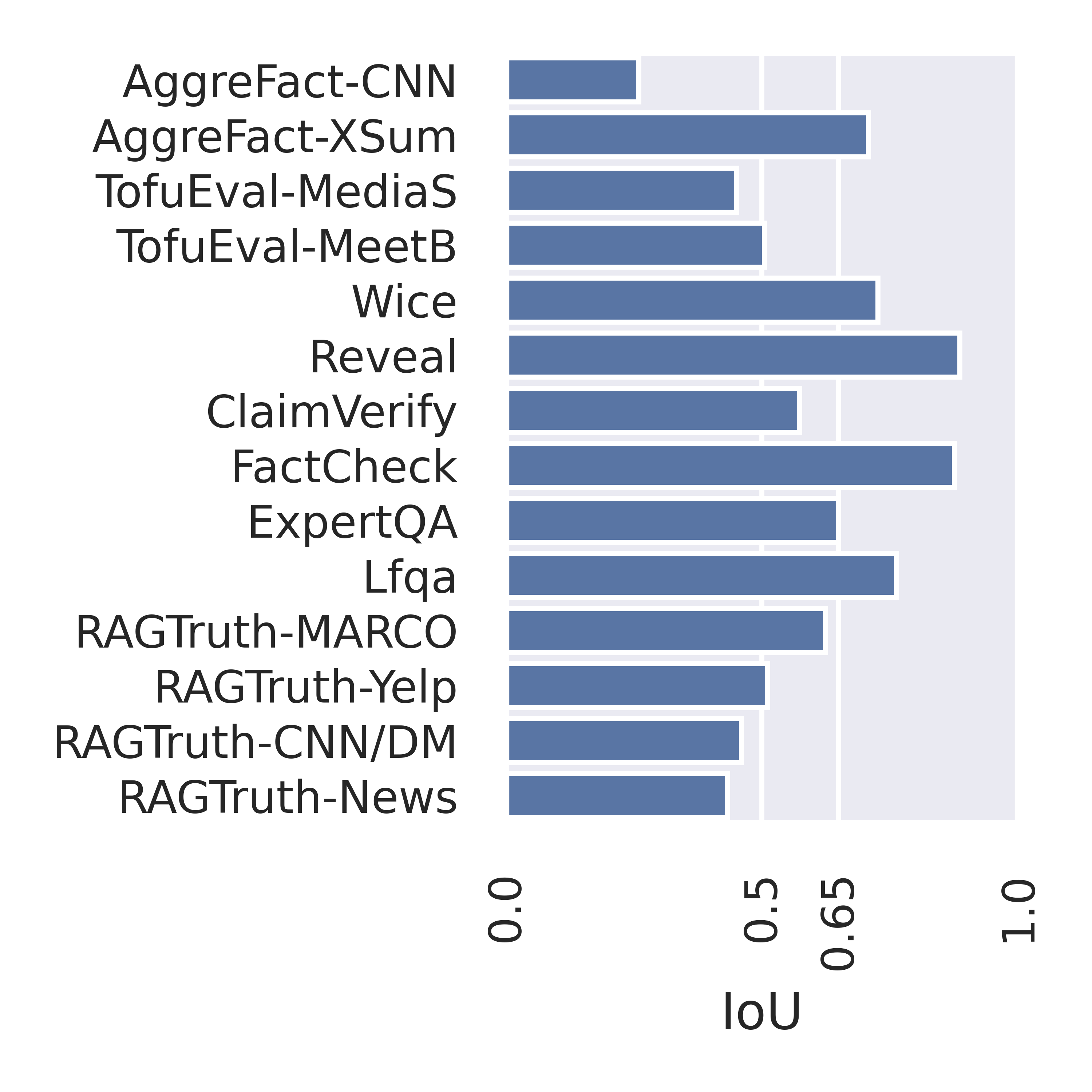}
    \vspace{-2em}
    \caption{\textbf{Intersection-over-Union of "unattributable" predictions by \gptf and \bespoke.} IoU less than 50\% on 5 of 14 datasets shows that the top-performing models (with very similar balanced accuracy of 76.2\% and 77.4\% respectively) have low consistency on what examples they predict as "unattributable".
    }
    \label{fig:prediction-iou-main}
\end{figure}

\begin{figure*}[tb]
    \centering
    \includegraphics[width=0.98\textwidth]{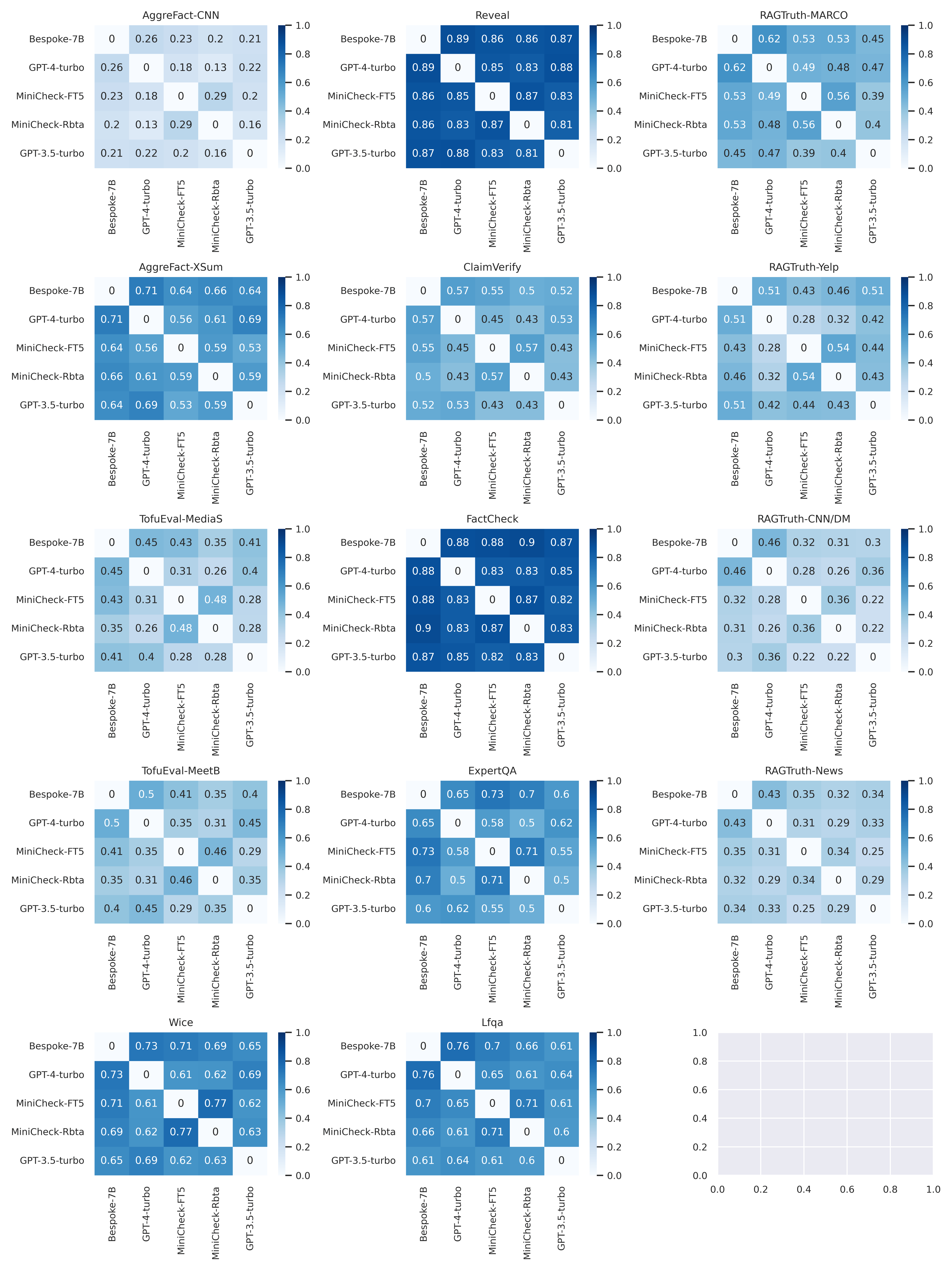}
    \caption{\textbf{Pairwise Intersection-over-Union of "unattributable" predictions by \ais metrics.}
    }
    \label{fig:prediction-iou-unattrib}
\end{figure*}

\begin{figure*}[tb]
    \centering
    \includegraphics[width=0.98\textwidth]{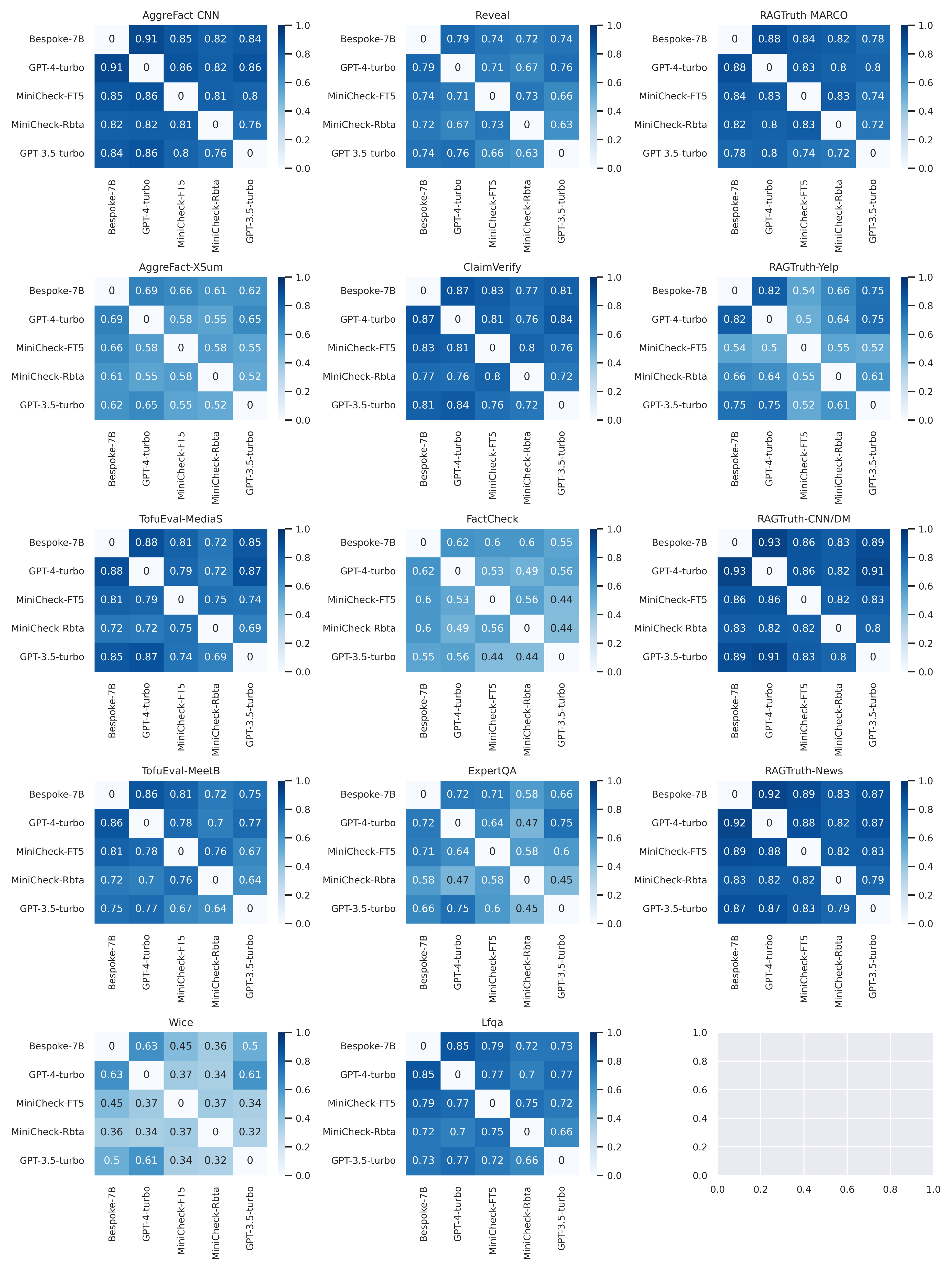}
    \caption{\textbf{Pairwise Intersection-over-Union of "attributable" predictions by \ais metrics.}
    }
    \label{fig:prediction-iou-attrib}
\end{figure*}

As discussed in \S~\ref{sec:f1-inconsistent}, Figure~\ref{fig:prediction-iou-main} demonstrates that the set of claims labeled as unattributable by two top-performing metrics \gptf and \bespoke has low overlap.
Figures~\ref{fig:prediction-iou-unattrib} and ~\ref{fig:prediction-iou-attrib} show the pairwise consistency (IoU) in predicting the label "attributable" and "unattributable" respectively between the different evaluation metrics on each dataset of \llmagg.

\end{document}